\documentclass{article} %
\usepackage{iclr2023_conference,times}

\usepackage{amsmath,amsfonts,bm}

\def\secref#1{section~\ref{#1}}

\def\eqref#1{equation~\ref{#1}}

\def\1{\bm{1}}

\def\vs{{\bm{s}}}

\DeclareMathAlphabet{\mathsfit}{\encodingdefault}{\sfdefault}{m}{sl}
\SetMathAlphabet{\mathsfit}{bold}{\encodingdefault}{\sfdefault}{bx}{n}

\usepackage{hyperref}
\usepackage{url}

\title{MOAT: Alternating Mobile Convolution and Attention Brings Strong Vision Models}

\author{
\centerline{Chenglin Yang$^1$\thanks{Work done while an intern at Google.}~~, Siyuan Qiao$^2$, Qihang Yu$^1$, Xiaoding Yuan$^1$, Yukun Zhu$^2$,}\\
\centerline{\textbf{Alan Yuille$^1$, Hartwig Adam$^2$, Liang-Chieh Chen$^2$}}\\
\centerline{$^1$The Johns Hopkins University}\\
\centerline{$^2$Google Research}
}

\definecolor{darkblue}{rgb}{0,0.08,0.45}
\hypersetup{
    colorlinks=true,
    citecolor=darkblue
}
\usepackage{url}            %
\usepackage{booktabs}       %
\usepackage{amsfonts}       %
\usepackage{nicefrac}       %
\usepackage{microtype}      %
\usepackage{xcolor}         %

\usepackage{graphicx}
\usepackage{amsmath}
\usepackage{amssymb}
\usepackage{multirow}  
\usepackage{color}
\usepackage{wrapfig}
\usepackage{makecell}
\usepackage[font=small]{caption}

\makeatletter
\usepackage{xspace}
\def\@onedot{\ifx\@let@token.\else.\null\fi\xspace}
\DeclareRobustCommand\onedot{\futurelet\@let@token\@onedot}

 \newcommand{\figureref}[1]{Fig\onedot~\ref{#1}}
\newcommand{\equref}[1]{Eq\onedot~\ref{#1}}
\newcommand{\tabref}[1]{Tab\onedot~\ref{#1}}

\def\eg{\emph{e.g}\onedot} 
\def\ie{\emph{i.e}\onedot} 
 
 \def\vs{\emph{vs}\onedot}
\def\wrt{w.r.t\onedot}

\newcommand{\name}{MOAT\xspace}
\definecolor{baselinecolor}{gray}{.9}
\newcommand{\baseline}[1]{\cellcolor{baselinecolor}{#1}}
\newcommand{\tinyname}{tiny-MOAT\xspace}

\usepackage{colortbl}

\iclrfinalcopy %
\begin{document}

\maketitle

\begin{abstract}
This paper presents MOAT, a family of neural networks that build on top of \textbf{MO}bile convolution (\ie, inverted residual blocks) and \textbf{AT}tention.
Unlike the current works that stack separate mobile convolution and transformer blocks, we effectively merge them into a MOAT block.
Starting with a standard Transformer block, we replace its multi-layer perceptron with a mobile convolution block, and further reorder it before the self-attention operation.
The mobile convolution block not only enhances the network representation capacity, but also produces better downsampled features.
Our conceptually simple MOAT networks are surprisingly effective, achieving 89.1\% / 81.5\% top-1 accuracy on ImageNet-1K / ImageNet-1K-V2 with ImageNet-22K pretraining.
Additionally, MOAT can be seamlessly applied to downstream tasks that require large resolution inputs by simply converting the global attention to window attention.
Thanks to the mobile convolution that effectively exchanges local information between pixels (and thus cross-windows), MOAT does not need the extra window-shifting mechanism.
As a result, on COCO object detection, MOAT achieves 59.2\% AP$^{\text{box}}$ with 227M model parameters (single-scale inference, and hard NMS), and on ADE20K semantic segmentation, MOAT attains 57.6\% mIoU with 496M model parameters (single-scale inference).
Finally, the tiny-MOAT family, obtained by simply reducing the channel sizes, also surprisingly outperforms several mobile-specific transformer-based models on ImageNet. The tiny-MOAT family is also benchmarked on downstream tasks, serving as a baseline for the community.
We hope our simple yet effective MOAT will inspire more seamless integration of convolution and self-attention.
Code is publicly available.\footnote{Official code in TensorFlow: \url{https://github.com/google-research/deeplab2}}
\end{abstract}

\section{Introduction}

The vision community has witnessed the prevalence of self-attention~\citep{bahdanau2014neural} and Transformers~\citep{vaswani2017attention}.
The success of Transformers in natural language processing motivates the creation of their variants for vision recognition.
The Vision Transformer (ViT)~\citep{dosovitskiy2020image} has great representation capacity with global receptive field.
However, it requires pretraining on a large-scale proprietary dataset~\citep{sun2017revisiting}.
Its unsatisfying performance, when trained with a small number of images, calls for the need of better training recipes~\citep{touvron2021training,steiner2021train} or architectural designs~\citep{liu2021swin,graham2021levit}.
On the other hand, ConvNet has been the dominant network choice since the advent of AlexNet~\citep{krizhevsky2012imagenet} in 2012.
Vision researchers have condensed the years of network design experience into multiple principles, and have started to incorporate them to vision transformers.
For example, there are some works adopting the ConvNet's hierarchical structure to extract multi-scale features for vision transformers~\citep{liu2021swin,fan2021multiscale,wang2022pvt}, and others proposing to integrate the translation equivariance of convolution into transformers~\citep{graham2021levit,d2021convit,xiao2021early}.

Along the same direction of combining the best from Transformers and ConvNets, CoAtNet~\citep{dai2021coatnet} and MobileViT~\citep{mehta2021mobilevit} demonstrate outstanding performance by stacking Mobile Convolution (MBConv) blocks (\ie, inverted residual blocks~\citep{sandler2018mobilenetv2}) and Transformer blocks (\ie, a self-attention layer and a Multi-Layer Perceptron (MLP)).
However, both works focus on the macro-level network design.
They consider MBConv and Transformer blocks as individual separate ones, and systematically study the effect of stacking them to strike a better balance between the remarkable efficiency of MBConv and strong capacity of Transformer.

In this work, on the contrary, we study the \textit{micro-level} building block design by taking a deeper look at the combination of MBConv and Transformer blocks.
We make two key observations after a careful examination of those blocks.
First, the MLP module in Transformer block is similar to MBConv, as both adopt the inverted bottleneck design.
However, MBConv is a more powerful operation by employing one extra $3\times3$ depthwise convolution (to encode local interaction between pixels), and more activation~\citep{hendrycks2016gaussian} and normalization~\citep{ioffe2015batch} are employed between convolutions. 
Second, to extract multi-scale features using Transformer blocks, one may apply the average-pooling (with stride 2) to input features before the self-attention layer.
However, the pooling operation reduces the representation capacity of self-attention.
Our observations motivate us to propose a novel \textbf{MO}bile convolution with \textbf{AT}tention (MOAT) block, which efficiently combines MBConv and Transformer blocks.
The proposed MOAT block modifies the Transformer block by first  replacing its MLP with a MBConv block, and then reversing the order of attention and MBConv.
The replacement of MLP with MBConv brings more representation capacity to the network, and reversing the order (MBConv comes before self-attention) delegates the downsampling duty to the strided depthwise convolution within the MBConv, learning a better downsampling kernel.

We further develop a family of MOAT models by stacking and increasing the channels of network blocks.
Surprisingly, our extremely \textit{simple} design results in a remarkable impact.
On the challenging ImageNet-1K classification benchmark~\citep{russakovsky2015imagenet}, our model (190M parameters) achieves 86.7\% top-1 accuracy without extra data.
When further pretraining on ImageNet-22K, our best model (483M parameters) attains 89.1\% / 81.5\% top-1 accuracy on ImageNet-1K (\tabref{tab:i1k21k_result}) / ImageNet-1K-V2 (\tabref{tab:imagevet_1k_v2}), setting a new state-of-the-art.

Additionally, MOAT can be \textit{seamlessly} deployed to downstream tasks that require large resolution inputs by simply converting the global attention to non-overlapping local window attention.
Thanks to the MBConv that effectively exchanges local information between pixels (enabling cross-window propagation), MOAT does not need the extra window-shifting mechanism~\citep{liu2021swin}.
As a result,
on COCO object detection~\citep{lin2014microsoft} and
ADE20K semantic segmentation~\citep{zhou2019semantic}, MOAT shows superior performances.
Specifically, on COCO object detection (\tabref{tab:coco_exp}), our best model (227M parameters), achieves 59.2\% AP$^{\text{box}}$ with single-scale inference and hard NMS, setting a new state-of-the-art in the regime of model size 200M with Cascade Mask R-CNN~\citep{cai2018cascade,he2017mask}.
On ADE20K semantic segmentation (\tabref{tab:ade20k_sem_exp}), our best model (496M parameters), adopting DeepLabv3+~\citep{deeplabv3plus2018}, attains 57.6\% mIoU with single-scale inference, also setting a new state-of-the-art in the regime of models using input size $641\times641$.

Finally, to explore the scalability of MOAT models, we \textit{simply} scale down the models by reducing the channel sizes (without any other change), resulting in the tiny-MOAT family, which also surprisingly outperforms mobile-specific transformer-based models, such as Mobile-Former~\citep{chen2022mobile} and MobileViTs~\citep{mehta2021mobilevit,mehta2022separable}.
Specifically, in the regime of model parameters 5M, 10M, and 20M, our tiny MOAT outperforms the concurrent MobileViTv2~\citep{mehta2022separable} by 1.1\%, 1.3\%, and 2.0\% top-1 accuracy on ImageNet-1K classification benchmark (\tabref{tab:i1k_result_tiny}). Furthermore, we benchmark \tinyname on COCO object detection and ADE20K semantic segmentation.

In summary, our method advocates the design principle of simplicity. Without inventing extra complicated operations, the proposed MOAT block effectively merges the strengths of both mobile convolution and self-attention into one block by a careful redesign. Despite its conceptual simplicity, impressive results have been obtained on multiple core vision recognition tasks. We hope our study will inspire future research on seamless integration of convolution and self-attention.

\section{Method}

Herein, we review the Mobile Convolution (MBConv)~\citep{sandler2018mobilenetv2} and Transformer~\citep{vaswani2017attention} blocks before introducing the proposed MOAT block.
We then present MOAT, a family of neural networks, targeting at different trade-offs between accuracy and model complexity.

\begin{figure}[!t]
  \centering
  \includegraphics[width=0.8\textwidth]{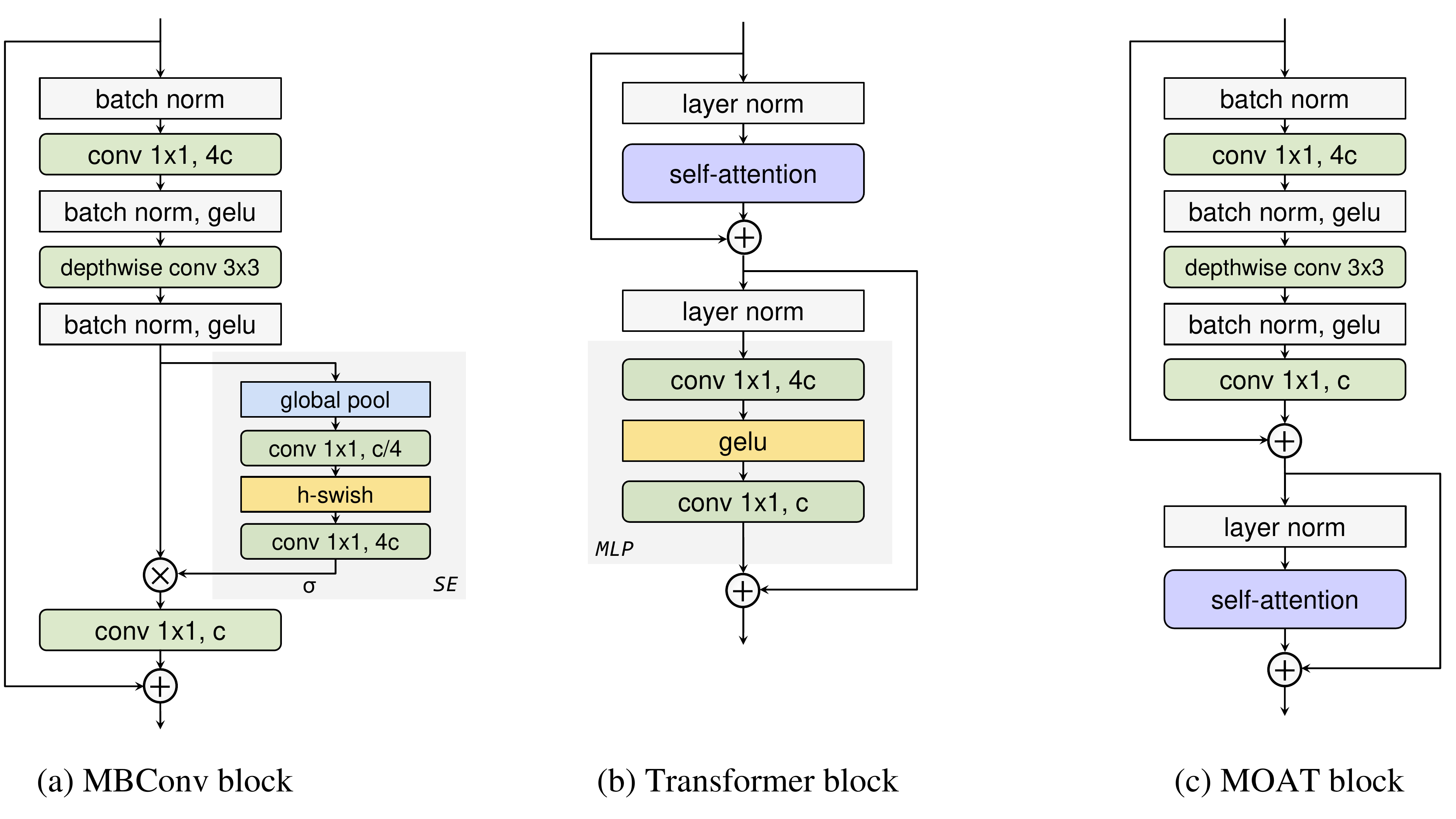}
  \caption{\textbf{Block comparison.} (a) The MBConv block~\citep{sandler2018mobilenetv2} employs the inverted bottleneck design with depthwise convolution and squeeze-and-excitation~\citep{hu2018squeeze} applied to the expanded features. (b) The Transformer block~\citep{vaswani2017attention} consists of a self-attention module and a MLP module. (c) The proposed MOAT block effectively combines them. %
  The illustration assumes the input tensor has channels $c$.
  }
  \label{fig: block_comparison}
\end{figure}

\subsection{Mobile Convolution and Transformer blocks}

\textbf{MBConv block.}
Also known as the inverted residual block, the Mobile Convolution (MBConv)~\citep{sandler2018mobilenetv2} block (\figureref{fig: block_comparison}~(a)) is an effective building block that has been widely used in mobile models~\citep{howard2019searching,mehta2021mobilevit} or efficient models~\citep{tan2019efficientnet,dai2021coatnet}. Unlike the bottleneck block in ResNet~\citep{he2016deep}, the MBConv block employs the design of an ``inverted bottleneck", together with the efficient depthwise convolution~\citep{howard2017mobilenets}.
Specifically, a $1\times1$ convolution is first applied to expand the input channels by a factor of 4. Then, a $3\times3$ depthwise  convolution is used to effectively capture the local spatial interactions between pixels. Finally, the features are projected back to the original channel size via a $1\times1$ convolution, enabling a residual connection~\citep{he2016deep}.
An optional Squeeze-and-Excitation (SE)~\citep{hu2018squeeze} module (which uses the global information to re-weight the channel activation) may also be used after the depthwise convolution, following MobileNetV3~\citep{howard2019searching}.
Note that one could tune the channel expansion ratio and depthwise convolution kernel size for better performance.
We fix them throughout the experiments for simplicity.

Formally, given an input tensor $x \in \mathbb{R}^{H \times W \times C}$ ($H, W, C$ are its height, width, and channels), the MBConv block is represented as follows:
\begin{align}
    \textbf{MBConv}(x) &= x + (\mathcal{N}_{2} \circ \mathcal{S} \circ \mathcal{D} \circ \mathcal{N}_{1}) (\text{BN} (x)),  \label{equ:mbcon}\\
    \mathcal{N}_{1}(x) &= \text{GeLU}(\text{BN}(\textbf{Conv}(x))), \label{equ:mbcon_N1}\\
    \mathcal{D}(x) &= \text{GeLU}(\text{BN}(\textbf{DepthConv}(x))), \label{equ:mbconv_D}\\
    \mathcal{S}(x) &= \sigma ( \text{MLP} ( \text{GAP} (x)) \cdot x, \label{equ:mbconv_S}\\
    \mathcal{N}_{2}(x) &= \textbf{Conv}(x),
    \label{equ:mbconv_N2}
\end{align}
where BN, GeLU, GAP, and MLP stand for Batch Normalization~\citep{ioffe2015batch}, Gaussian error Linear Unit~\citep{hendrycks2016gaussian}, Global Average Pooling, and Multi-Layer Perceptron (with reduction ratio 4 and hard-swish~\citep{ramachandran2017searching}), respectively. The MBConv block consists of four main functions: $\mathcal{N}_{1}$, $\mathcal{D}$, $\mathcal{S}$, and $\mathcal{N}_{2}$, which correspond to the $1\times1$ convolution for channel expansion (by $4\times$), $3\times3$ depthwise convolution, squeeze-and-excitation~\citep{hu2018squeeze} ($\sigma$ is the sigmoid function), and $1\times1$ convolution for channel projection (by $4\times$), respectively.

\textbf{Transformer block.}
The Transformer~\citep{vaswani2017attention} block (\figureref{fig: block_comparison}~(b)) is a powerful building block that effectively captures the global information via the data-dependent self-attention operation. It consists of two main operations: self-attention and MLP. The self-attention operation computes the attention map based on the pairwise similarity between every pair of pixels in the input tensor, thus enabling the model's receptive field to encompass the entire spatial domain. Additionally, the attention map dynamically depends on the input, enlarging the model's representation capacity (unlike the convolution kernels, which are data-independent). The MLP operation contains two $1\times1$ convolutions, where the first one expands the channels (by $4\times$), the second one shrinks back the channels, and GeLU non-linearity is used in-between.

Formally, given an input tensor $x \in \mathbb{R}^{H \times W \times C} $, the Transformer block is represented as follows: 
\begin{align}
    \textbf{Transformer}(x) &= x + (\mathcal{M}_{2} \circ \mathcal{M}_{1} \circ \textbf{Attn}) (\text{LN}(x)),  \\
    \mathcal{M}_{1}(x) &= \text{GeLU}(\textbf{Conv}(\text{LN}(x))), \\
    \mathcal{M}_{2}(x) &= \textbf{Conv}(x),
    \label{equ: tfm}
\end{align}
where LN and Attn denote the Layer Normalization~\citep{ba2016layer}, and self-attention~\citep{vaswani2017attention}. The self-attention operation also includes a residual connection~\citep{he2016deep}, which is not shown in the equations for simplicity, while the MLP operation is represented by two functions $\mathcal{M}_{1}$ and $\mathcal{M}_{2}$, which correspond to the $1\times1$ convolution for channel expansion (by $4\times$) and $1\times1$ convolution for channel projection, respectively.

\subsection{MObile convolution with ATtention (MOAT) block}

\textbf{Comparing MBConv and Transformer blocks.}
Before getting into the architecture of our MOAT block, it is worthwhile to compare the MBConv~\citep{sandler2018mobilenetv2} and Transformer~\citep{vaswani2017attention} blocks, which helps to understand our design motivations. Specifically, we make the following key observations.

First, both MBConv and Transformer blocks advocate the ``inverted bottleneck" design, where the channels of input tensors are expanded and then projected by $1\times1$ convolutions. However, MBConv additionally employs a $3\times3$ depthwise convolution between those two $1\times1$ convolutions, and there are both batch normalization and GeLU activation between the convolutions.

Second, to capture the global information, the MBConv block may employ a Squeeze-and-Excitation (SE) module, while the Transformer block adopts the self-attention operation. Note that the SE module squeezes the spatial information via a global average pooling, while the self-attention module maintains the tensor's spatial resolution. 

Third, the downsampling operation is performed at different places within the block. To downsample the features, the standard MBConv block uses the strided depthwise convolution, while the Transformer block, deployed in the modern hybrid model CoAtNet~\citep{dai2021coatnet}, adopts an average-pooling operation before the self-attention.

\textbf{MOAT block.}
Given the above observations, we now attempt to design a new block that effectively merges the best from both MBConv and Transformer blocks. We begin with the powerful Transformer block, and gradually refine over it.

Based on the first observation, both MBConv and Transformer blocks employ the ``inverted bottleneck" design.
Since depthwise convolution could effectively encode local interaction between pixels, which is crucial for modeling the translation equivariance in ConvNets, we thus start to add the depthwise convolution to Transformer's MLP module.
However, we did not observe any performance improvement until we also added the extra normalization and activations between convolutions.%

For the second observation, we simply do not add the SE module to the MBConv block. The self-attention operation is kept to capture the global information.

We found the third observation critical. The downsampling operation (average-pooling) right before the self-attention operation in Transformer block slightly reduces its representation capacity. On the other hand, the MBConv block is well-designed for the downsampling operation with the strided depthwise convolution, which effectively learns the downsampling convolution kernel for each input channel. Therefore, we further reorder the ``inverted bottleneck" (containing depthwise convolution) before the self-attention operation, delegating the downsampling operation to depthwise convolution. In this way, we need no extra downsampling layer like average-pooling in CoAtNet~\citep{dai2021coatnet}, or patch-embedding layers in Swin~\citep{liu2021swin} and ConvNeXt~\citep{liu2022convnet}.  Finally, it results in our \textbf{MO}bile convolution with \textbf{AT}tention (MOAT) block, as illustrated in~\figureref{fig: block_comparison}~(c).

Formally, given an input tensor $x \in \mathbb{R}^{H \times W \times C} $, the MOAT block is represented as follows:
\begin{align}
    \textbf{MOAT}(x) &= x + (\textbf{Attn} \circ \mathcal{N}_{2} \circ \mathcal{D} \circ \mathcal{N}_{1}) (\text{BN} (x)),
    \label{equ: moat}
\end{align}
where MBConv (w/o SE) contains functions $\mathcal{N}_{1}$ (\equref{equ:mbcon_N1}), $\mathcal{D}$ (\equref{equ:mbconv_D}), and $\mathcal{N}_{2}$ (\equref{equ:mbconv_N2}), and Attn denotes the self-attention operation. The MOAT block then simply consists of MBConv (w/o SE) and the self-attention operation, successfully combining the best from the MBConv block and Transformer block into one (which we will show empirically).

\subsection{Meta Architecture}

\textbf{Macro-level network design.}
After developing the MOAT block, we then study how to effectively stack them to form our base model.
We adopt the same strategy as the existing works~\citep{liu2021swin,wang2021pyramid,graham2021levit,xiao2021early,dai2021coatnet,mehta2021mobilevit}.
Specifically, we summarize several key findings from those works, and use them as design principles of our meta architecture.

\begin{itemize}
    \item Employing convolutions in the early stages improves the performance and training convergence of %
    Transformer models~\citep{wu2021cvt,graham2021levit,xiao2021early}.
    \item The Mobile Convolution (MBConv)~\citep{sandler2018mobilenetv2} blocks are also effective building blocks in the hybrid Conv-Transformer models~\citep{dai2021coatnet,mehta2021mobilevit}.
    \item Extracting multi-scale backbone features benefits the downstream tasks, such as detection and segmentation    ~\citep{liu2021swin,wang2021pyramid,fan2021multiscale,heo2021rethinking}.
\end{itemize}

As a result, our meta architecture consists of the convolutional stem, MBConv blocks, and MOAT blocks.
Additionally, through the ablation study in the appendix, we found the layer layout proposed by CoAtNet-1~\citep{dai2021coatnet} effective.
We thus follow their layer layout, resulting in our base model MOAT-1.
To form the MOAT model family, we then scale down or up MOAT-1 in the dimensions of number of blocks and number of channels, as shown in~\tabref{tab:moat_family}.
We only scale the number of blocks in the third and fourth stages (out of five stages).
The downsampling operation is performed in the first block of each stage.
Note that our base model MOAT-1 and CoAtNet-1 share the same layer layout and channel sizes. However, we take a different scaling strategy: our MOAT is scaled up (or down) by alternatively increasing the depth and expanding the width between variants.

\begin{table}[!ht]
    \centering
    \caption{\textbf{MOAT variants} differ in the number of blocks \texttt{B} and number of channels \texttt{C} in each stage.
    }
    \scalebox{0.72}{
    \begin{tabular}{l c | cc | cc | cc | cc | cc | c cccc}
    \multirow{2}{*}{block} & \multirow{2}{*}{stride}
    & \multicolumn{2}{c|}{\name-0} 
    & \multicolumn{2}{c|}{\name-1} 
    & \multicolumn{2}{c|}{\name-2} 
    & \multicolumn{2}{c|}{\name-3} 
    & \multicolumn{2}{c|}{\name-4}
    & \multicolumn{5}{c}{\tinyname-\{0,1,2,3\}}
    \\
    {} & {} 
    & \texttt{B} & \texttt{C} 
    & \texttt{B} & \texttt{C}  
    & \texttt{B} & \texttt{C}   
    & \texttt{B} & \texttt{C}   
    & \texttt{B} & \texttt{C}
    & \texttt{B} & \multicolumn{4}{c}{\texttt{C}}  
    \\
    \toprule
    conv & 2
        & \texttt{2} & \texttt{64}
        & \texttt{2} & \texttt{64}
        & \texttt{2} & \texttt{128}
        & \texttt{2} & \texttt{160}
        & \texttt{2} & \texttt{256}
        & \texttt{2} & \texttt{32} & \texttt{40} & \texttt{56} & \texttt{80}
        \\
    MBConv & 4
        & \texttt{2} & \texttt{96}
        & \texttt{2} & \texttt{96}
        & \texttt{2} & \texttt{128} 
        & \texttt{2} & \texttt{160}
        & \texttt{2} & \texttt{256}
        & \texttt{2} & \texttt{32} & \texttt{40} & \texttt{56} & \texttt{80}
        \\
    MBConv & 8
        & \texttt{3} & \texttt{192}
        & \texttt{6} & \texttt{192}
        & \texttt{6} & \texttt{256}
        & \texttt{12} & \texttt{320} 
        & \texttt{12} & \texttt{512}
        & \texttt{3} & \texttt{64} & \texttt{80} & \texttt{112} & \texttt{160}
        \\
    MOAT & 16
        & \texttt{7}  & \texttt{384}
        & \texttt{14} & \texttt{384}
        & \texttt{14} & \texttt{512}
        & \texttt{28} & \texttt{640}
        & \texttt{28} & \texttt{1024} & \texttt{7} & \texttt{128} & \texttt{160} & \texttt{224} & \texttt{320}
        \\
    MOAT & 32
        & \texttt{2} & \texttt{768}
        & \texttt{2} & \texttt{768}
        & \texttt{2} & \texttt{1024}
        & \texttt{2} & \texttt{1280}
        & \texttt{2} & \texttt{2048} & \texttt{2} & \texttt{256} & \texttt{320} & \texttt{448} & \texttt{640}
        \\
    \end{tabular}
    }
    \label{tab:moat_family}
\end{table}

\section{Experimental Results}
\label{sec: experimental_results}

In this section, we show that \name variants are effective on the  ImageNet-1K~\citep{russakovsky2015imagenet} image classification.
We then deploy them to other recognition tasks, including COCO object detection~\citep{lin2014microsoft}, instance segmentation~\citep{hariharan2014simultaneous},
and ADE20K~\citep{zhou2019semantic} semantic segmentation.
MOAT can be seamlessly applied to downstream tasks.
For small resolution inputs, we directly fine-tune the global attention, while for large resolution inputs, we simply convert the global attention to non-overlapping local window attention without using extra window-shifting mechanism.
The detailed experiment setup could be found in the appendix.

\begin{table}[!t]
\small
    \centering
    \caption{\textbf{Performance on ImageNet-1K}. \textbf{1K only:} Using ImageNet-1K only. \textbf{22K + 1K:} ImageNet-22K pretraining and ImageNet-1K fine-tuning. \tabref{tab:i1k21k_result_full} shows comparisions with more SOTA methods and \tabref{tab:imagevet_1k_v2} reports the performances on ImageNet-1K-V2.
    }
    \scalebox{0.8}{\begin{tabular}{c l c c c c c}
        \multicolumn{2}{c}{model} & eval size & params & FLOPs &  \multicolumn{2}{c}{ImageNet-1K top-1 accuracy} \\
        \toprule
        \multicolumn{5}{c}{} & 1K only & 22K+1K \vspace{0.5em} \\
        \multirow{7}{*}{ConvNets}
        & EfficientNetV2-L~\citep{tan2021efficientnetv2} & $480^2$ & 120M & 53B & 85.7 & - \\
        & EfficientNetV2-XL~\citep{tan2021efficientnetv2} & $480^2$ & 208M & 94B & - & 87.3 \\
        & ConvNeXt-T~\citep{liu2022convnet} & $224^2$ & 29M & 4.5B & 82.1 & 82.9 \\
        & ConvNeXt-L~\citep{liu2022convnet} & $384^2$ & 198M & 101.0B & 85.5 & 87.5 \\
        & ConvNeXt-XL~\citep{liu2022convnet} & $384^2$ & 350M & 179.0B & - & 87.8 \\
        \cmidrule(lr){1-7}%
        \multirow{4}{*}{ViTs}
        & PVT-Large~\citep{wang2021pyramid}  & $224^2$ & 61.4M  & 9.8B & 81.7 & - \\
        & Swin-T~\citep{liu2021swin}  & $224^2$ & 28M  & 4.5B & 81.3  & - \\
        & Swin-L~\citep{liu2021swin}  & $384^2$ & 197M & 103.9B & -  & 87.3 \\
        & SwinV2-L~\citep{liu2021swin}  & $384^2$ & 197M & 115.4B & - & 87.7 \\
        & MViTv2-H~\citep{li2022improved}  & $512^2$ & 667M & 763.5B & - & 88.8 \\
        \cmidrule(lr){1-7}%
        \multirow{4}{*}{Hybrid} 
        & PVTv2-B5~\citep{wang2022pvt} & $224^2$ & 82M &  11.8B & 83.8 & - \\
        & MaxViT-XL~\citep{tu2022maxvit}  & $512^2$ & 475M & 535.2B & - & 88.7 \\
        & CoAtNet-0~\citep{dai2021coatnet} & $224^2$ & 25M  & 4.2B & 81.6 & - \\
        & CoAtNet-3~\citep{dai2021coatnet} & $384^2$ & 168M & 107.4B & 85.8 & 87.6\\
        & CoAtNet-4~\citep{dai2021coatnet} & $512^2$ & 275M & 360.9B & - & 88.6 \\
        \midrule \midrule
        \multirow{12}{*}{\makecell{\bf Hybrid \\ \bf (ours)}} 
        & \name-0 & $224^2$ & 27.8M  & 5.7B & 83.3 & 83.6 \\
        & \name-1 & $224^2$ & 41.6M  & 9.1B & 84.2 & 84.9  \\
        & \name-2 & $224^2$ & 73.4M & 17.2B & 84.7 & 86.0 \\
        & \name-3 & $224^2$ & 190.0M & 44.9B & 85.3 & 86.8 \\
        \cmidrule(lr){2-7}%
        & \name-0 & $384^2$ & 27.8M & 18.2B & 84.6  & 85.7 \\
        & \name-1 & $384^2$ & 41.6M  & 29.6B & 85.9  & 87.0 \\
        & \name-2 & $384^2$ & 73.4M  & 54.3B & 86.2 & 87.5 \\
        & \name-3 & $384^2$ & 190.0M & 141.2B & 86.5 & 88.2 \\
        \cmidrule(lr){2-7}%
        & \name-1 & $512^2$ & 41.6M & 58.7B & 86.2 & 87.2 \\
        & \name-2 & $512^2$ & 73.4M & 104.6B  & 86.5 & 87.7 \\
        & \name-3 & $512^2$ & 190.0M & 271.0B & 86.7 &  88.4 \\
        \cmidrule(lr){2-7}%
        & \name-4 & $512^2$ & 483.2M & 648.5B & - &  89.1 \\
    \end{tabular}}
    \label{tab:i1k21k_result}
\end{table}

\begin{center}
\begin{minipage}[t]{0.48\linewidth}
\includegraphics[width=\linewidth]{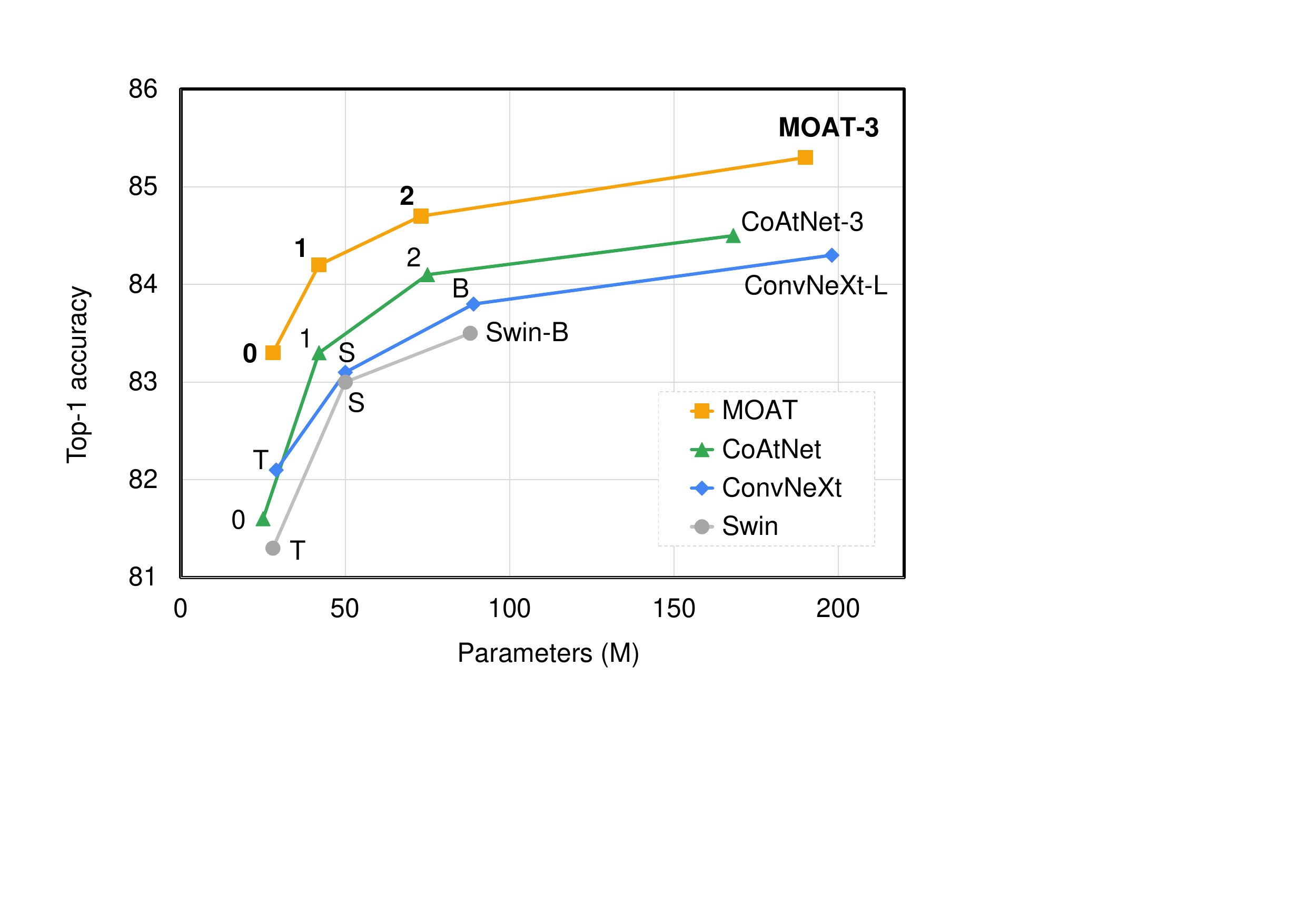}
\captionof{figure}{\textbf{Parameters \vs accuracy} using ImageNet-1K only with input size 224.
}
\label{fig: 1k_224_params}
\end{minipage}
\hfill
\begin{minipage}[t]{0.48\linewidth}
\includegraphics[width=\linewidth]{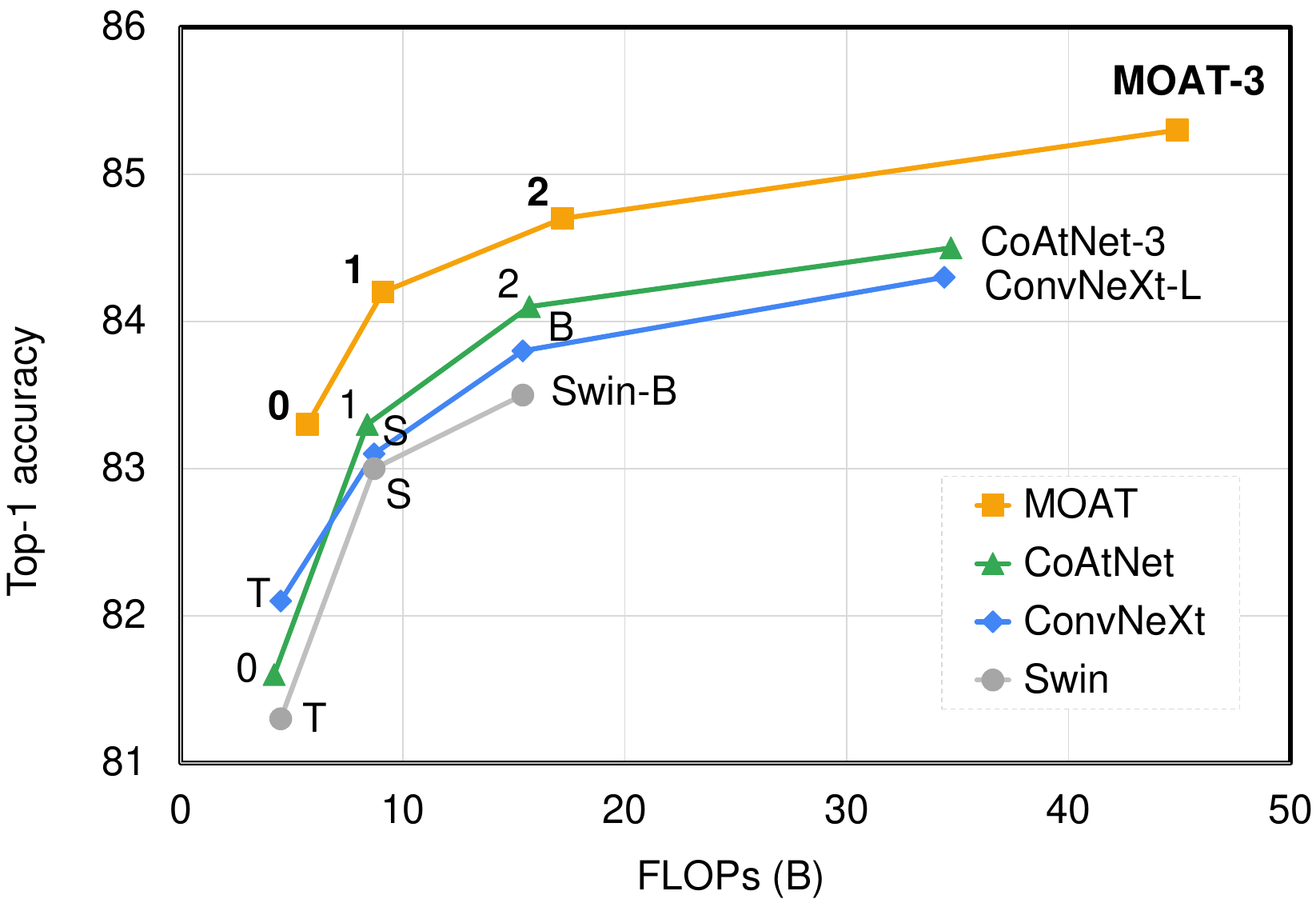}
\captionof{figure}{\textbf{FLOPs \vs accuracy} using ImageNet-1K only with input size 224.}
\label{fig: 1k_224_flops}
\end{minipage}
\begin{minipage}[t]{0.47\linewidth}
\includegraphics[width=\linewidth]{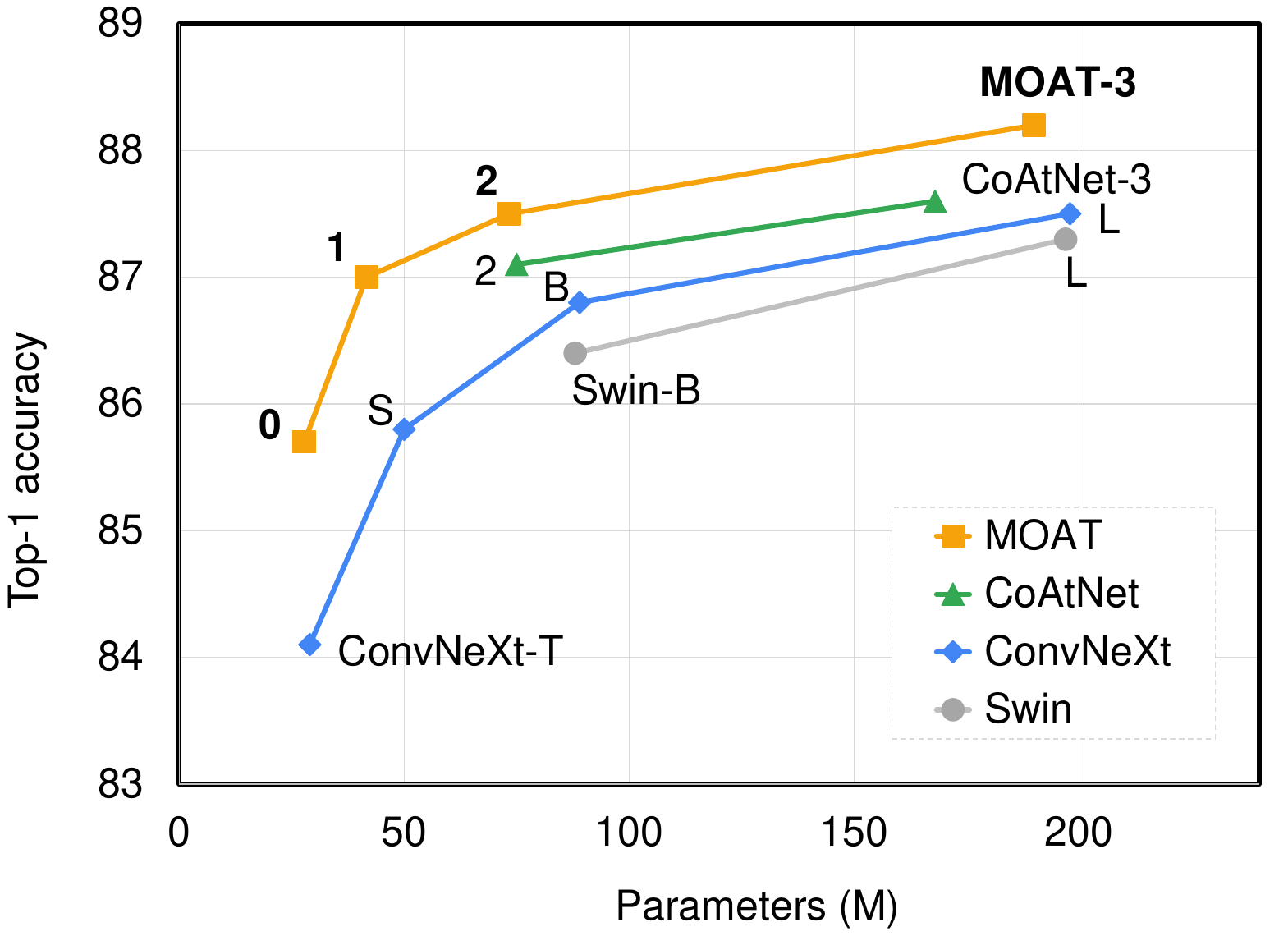}
\captionof{figure}{\textbf{Parameters \vs accuracy} using ImageNet-22K and ImageNet-1K with input size 384.}
\label{fig: 22k_384_params}
\end{minipage}
\hfill
\begin{minipage}[t]{0.49\linewidth}
\includegraphics[width=\linewidth]{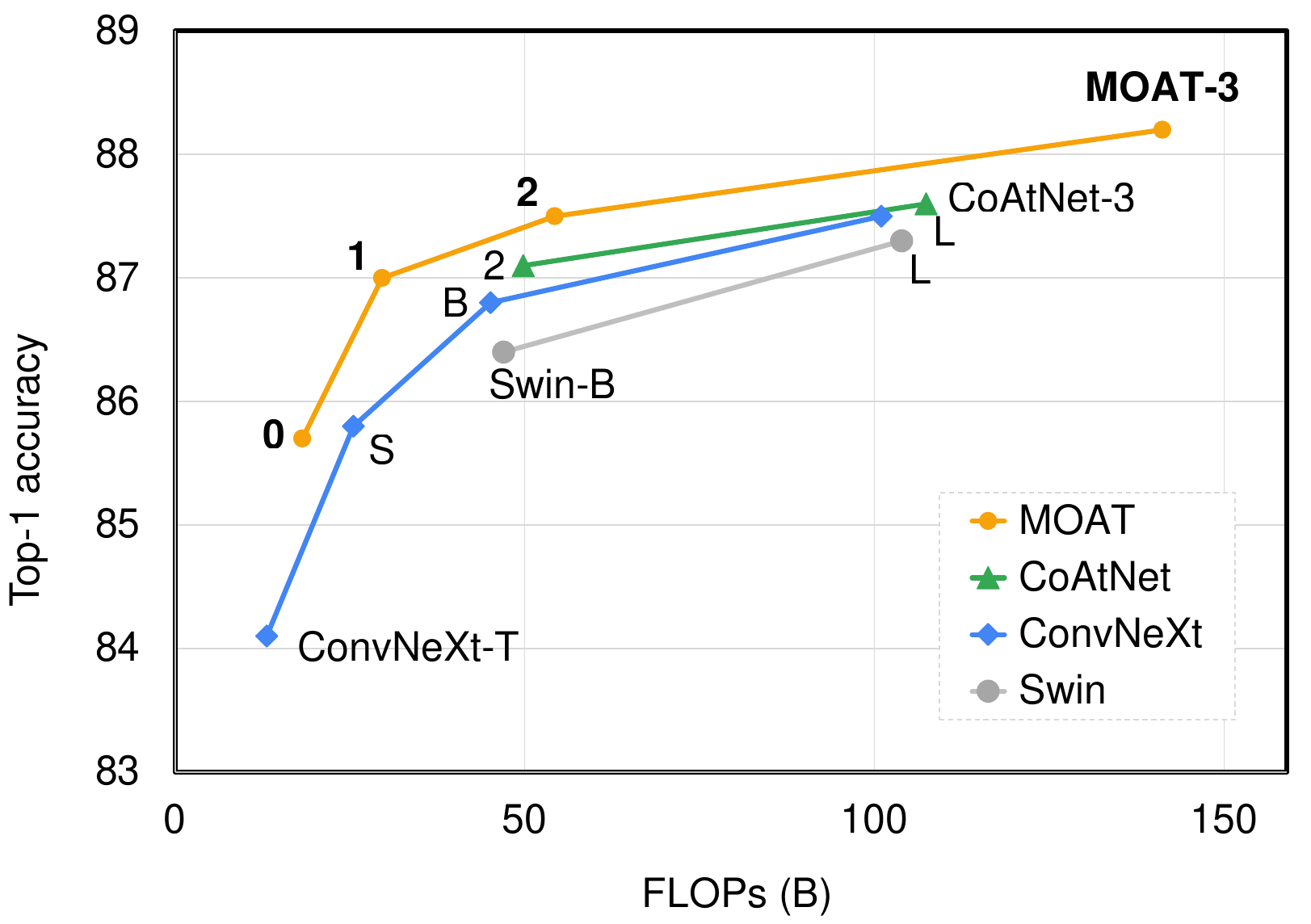}
\captionof{figure}{\textbf{FLOPs \vs accuracy} using ImageNet-22K and ImageNet-1K with input size 384.}
\label{fig: 22k_384_flops}
\end{minipage}
\end{center}

\textbf{ImageNet Image Classification.}
In~\tabref{tab:i1k21k_result}, we include the current state-of-art methods in the categories of ConvNets, ViTs and Hybrid models. 
At similar model costs (parameters or FLOPs), our \name models consistently outperform all of them.
Specifically, with the ImageNet-1K data only and input size 224, for light-weight models, our \name-0 significantly outperforms ConvNeXt-T~\citep{liu2022convnet},
Swin-T~\citep{liu2022convnet}, and CoAtNet-0~\citep{dai2021coatnet} by 1.2\%, 2.0\%, and 1.7\%, respectively. For large-scale models using input size 384, \name-3 is able to surpass ConvNeXt-L, CoAtNet-3 by 1.0\% and 0.7\%, respectively. 
With the ImageNet-22K pretraining and input size 384, the prior arts ConvNeXt-L, Swin-L, and CoAtNet-3 already show strong performances (87.5\%, 87.3\% and 87.6\%), while our \name-3 achieves the score of 88.2\%, outperforming them by 0.7\%, 0.9\%, and 0.6\%, respectively.
For ImageNet-1K and input size 224, we plot the performances \vs parameters and FLOPs in~\figureref{fig: 1k_224_params} and ~\figureref{fig: 1k_224_flops}, respectively.
For ImageNet-22K pretraining and input size 384, we plot the performances \vs parameters and FLOPs in ~\figureref{fig: 22k_384_params} and ~\figureref{fig: 22k_384_flops}, respectively.
In the figures, \name clearly demonstrates the best performance in all computation regimes.
Finally, our largest model \name-4, with ImageNet-22K and input size 512, further attains 89.1\% accuracy. %

\textbf{COCO Detection.}
\tabref{tab:coco_exp} summarizes the COCO object detection (box) and instance segmentation (mask) results.
Our \name backbones significantly outperform the baseline methods, including Swin~\citep{liu2021swin} and ConvNeXt~\citep{liu2022convnet} across different model sizes.
Specifically, our \name-0 outperforms Swin-T and ConvNeXt-T by 5.4\% and 5.5\% AP$^{\text{box}}$ (3.7\% and 3.7\% AP$^{\text{mask}}$).
Our \name-1 surpasses Swin-S and ConvNeXt-S by 5.9\% and 5.8\% AP$^{\text{box}}$ (4.3\% and 4.0\% AP$^{\text{mask}}$).
Our \name-2, with 110M parameters, is still 5.5\% and 4.5\% AP$^{\text{box}}$ (3.5\% and 2.4\% AP$^{\text{mask}}$) better than Swin-B and ConvNeXt-B.
Finally, our \name-3, using 227M parameters, achieves 59.2\% AP$^{\text{box}}$ (50.3\% AP$^{\text{mask}}$), setting a new state-of-the-art in the regime of model size 200M that is built on top of Cascade Mask R-CNN~\citep{cai2018cascade,he2017mask}.
More comparisons with smaller input size
can be found in~\tabref{tab: more_coco_exp}. \textbf{For \tinyname}, \tinyname-0/1 achieve the same performance as Swin-T/S and ConvNeXt-T/S but only use less than half of the parameters. Furthermore, \tinyname-3 is pretrained with ImageNet-1K and attains 55.2 AP$^{\text{box}}$ with 57M parameters, surpassing the ImageNet-22k pretrained Swin-L (53.9 AP$^{\text{box}}$ with 254M parameters) and ConvNeXt-L (54.8 AP$^{\text{box}}$ with 255M parameters).

\begin{table}[!ht]
    \centering
    \caption{\textbf{Object detection and instance segmentation on the COCO 2017 \texttt{val} set}.
    We employ Cascade Mask-RCNN, and single-scale inference (hard NMS). \dag: use ImageNet-22K pretrained weights. When using \tinyname series as backbones, most of the model parameters come from the decoder. More comparisons at input size 896 is reported in~\tabref{tab: more_coco_exp}.
    }
    \label{tab:coco_exp}
    \scalebox{0.76}{
    \begin{tabular}{l|ccc|ccc|ccc}
    backbone & input size & params & FLOPs & AP$^{\text{box}}$ & AP$^{\text{box}}_{\text{50}}$ & AP$^{\text{box}}_{\text{75}}$ & AP$^{\text{mask}}$ & AP$^{\text{mask}}_{\text{50}}$ & AP$^{\text{mask}}_{\text{75}}$ \\
    \toprule
    Swin-T & $1280\times800$ & 86M & 745B & 50.5 & 69.3 & 54.9 & 43.7 & 66.6 & 47.1 \\
    Swin-S & $1280\times800$ & 107M & 838B & 51.8 & 70.4 & 56.3 & 44.7 & 67.9 & 48.5 \\
    Swin-B\dag & $1280\times800$ & 145M & 982B & 53.0 & 71.8 & 57.5 & 45.8 & 69.4 & 49.7\\
    Swin-L\dag & $1280\times800$ & 254M & 1382B & 53.9 & 72.4 & 58.8 & 46.7 & 70.1 & 50.8 \\
    \midrule
    ConvNeXt-T & $1280\times800$ & 86M & 741B & 50.4 & 69.1 & 54.8 & 43.7 & 66.5 & 47.3 \\
    ConvNeXt-S & $1280\times800$ & 108M & 827B & 51.9 & 70.8 & 56.5 & 45.0 & 68.4 & 49.1 \\
    ConvNeXt-B\dag & $1280\times800$ & 146M & 964B & 54.0 & 73.1 & 58.8 & 46.9 & 70.6 & 51.3\\
    ConvNeXt-L\dag & $1280\times800$ & 255M & 1354B & 54.8 & 73.8 & 59.8 & 47.6 & 71.3 & 51.7\\
    \midrule \midrule
    \tinyname-0 & $1344\times1344$ & 41M & 612B & 50.5 & 69.3 & 56.0 & 43.3 & 66.6 & 47.3 \\
    \tinyname-1 & $1344\times1344$ & 42M & 628B & 51.9 & 71.6 & 56.1 & 44.6 & 68.4 & 48.3 \\
    \tinyname-2 & $1344\times1344$ & 47M & 669B & 53.0 & 72.2 & 58.0 & 45.0 & 69.4 & 48.8 \\
    \tinyname-3 & $1344\times1344$ & 57M & 754B & 55.2 & 74.8 & 60.6 & 47.0 & 71.8 & 51.2 \\
    \midrule
    \name-0 & $1344\times1344$ & 65M & 799B & 55.9 & 73.9 & 60.9 & 47.4 & 70.9 & 52.1 \\
    \name-1 & $1344\times1344$ & 79M & 921B & 57.7 & 76.0 & 63.4 & 49.0 & 73.4 & 53.2 \\
    \name-2\dag & $1344\times1344$ & 110M & 1217B & 58.5 & 76.6 & 64.3 & 49.3 & 73.9 & 53.9 \\
    \name-3\dag & $1344\times1344$ & 227M & 2216B & 59.2 & 77.8 & 64.9 & 50.3 & 74.8 & 55.5 \\
    \end{tabular}
    }
\end{table}

\textbf{ADE20K Semantic Segmentation.}
In~\tabref{tab:ade20k_sem_exp}, when using input size $513^2$,
\name consistently outperforms the ConvNeXt counterparts. \name-0 surpasses ConvNeXt-T by 3.0\% mIoU. Moreover, \name-2, with ImageNet-22k pretraining, surpasses ConvNeXt-B by 3.1\%. The larger \name-3 and \name-4 further outperform ConvNeXt-L and ConvNeXt-XL by 4.9\% and 5.4\%, respectively.
Finally, when using input size $641^2$, our \name-4 achieves the performance of 57.6\% mIoU, setting a new state-of-the-art in the regime of models using input size $641^2$.
\textbf{For \tinyname}, \tinyname-3 achieves comparable performance with ConvNeXt-S with less than half of the parameters.

\begin{table}[!ht]
\small
    \centering
    \caption{\textbf{Semantic segmentation on ADE20K \texttt{val} set.}
    We employ DeepLabv3+ (\textbf{single-scale} inference).
    Results for ConvNeXt and \name are obtained using the official code-base~\citep{deeplab2_2021} with the same training recipe.
    \dag: use ImageNet-22K pretrained weights.
    }
    \label{tab:ade20k_sem_exp}
    \begin{minipage}{0.49\linewidth}{
        \begin{center}
            \scalebox{0.76}{\begin{tabular}{l c c c c}
            backbone & input & params & FLOPs & mIoU (\%) \\
            \toprule
            ConvNeXt-T & $513^2$ & 34.2M & 47.6B & 45.8 \\
            ConvNeXt-S & $513^2$ & 55.8M & 70.8B & 47.8  \\
            ConvNeXt-B\dag & $513^2$ & 95.8M & 119.5B & 50.5 \\
            ConvNeXt-L\dag & $513^2$ & 208.3M & 256.4B & 51.0 \\
            ConvNeXt-XL\dag & $513^2$ & 364.0M & 446.2B & 51.8 \\
            \end{tabular}}
        \end{center}}
    \end{minipage}
    \begin{minipage}{0.49\linewidth}{
        \begin{center}
            \scalebox{0.76}{\begin{tabular}{l c c c c}
            backbone & input & params & FLOPs & mIoU (\%) \\
            \toprule
                \tinyname-0  & $513^2$ & 5.6M & 11.8B & 41.2 \\
                \tinyname-1  & $513^2$ & 7.8M & 15.2B & 43.1 \\
                \tinyname-2  & $513^2$ & 13.2M & 23.8B & 44.9 \\
                \tinyname-3  & $513^2$ & 24.2M & 41.2B & 47.5 \\
                \midrule
                \name-0  & $513^2$ & 33.3M & 61.3B & 48.8 \\
                \name-1  & $513^2$ & 47.0M & 85.4B & 51.8 \\
                \name-2\dag  & $513^2$ & 80.5M & 144.3B & 53.6 \\
                \name-3\dag  & $513^2$ & 198.4M & 331.5B & 55.9 \\
                \name-4\dag  & $513^2$ & 496.3M & 779.9B & 57.2 \\
                \midrule
                \name-2\dag  & $641^2$ & 80.5M & 242.0B & 54.7 \\
                \name-3\dag  & $641^2$ & 198.4M & 554.7B & 56.5 \\
                \name-4\dag  & $641^2$ & 496.3M & 1273.5B & 57.6 \\
            \end{tabular}}
        \end{center}}
    \end{minipage}
\end{table}

\textbf{tiny-MOAT on ImageNet.}
We simply scale down the channels of MOAT-0 to obtain the \tinyname family without any specific adaptions.
In the left of~\tabref{tab:i1k_result_tiny},
with the similar model parameters, \tinyname-0/1/2 surpass the Mobile-Former counterparts by 6.8\%, 5.5\%, and 4.3\%, respectively.
In the right of~\tabref{tab:i1k_result_tiny},
our \tinyname also shows stronger performances than MobileViT~\citep{mehta2021mobilevit}.
Even compared with the concurrent work MobileViTv2~\citep{mehta2022separable}, \tinyname-1/2/3 surpass their counterparts by 1.1\%, 1.3\%, and 2.1\%, respectively.

\begin{table}[!ht]
\small
    \centering
     \caption{\textbf{Performances of \tinyname family on ImageNet-1K.}}
    \begin{minipage}{0.49\linewidth}{
        \begin{center}
            \scalebox{0.84}{\begin{tabular}{l c c c}
            input size $224^{2}$ & params & FLOPs & \multicolumn{1}{c}{top-1 acc.} \\
            \toprule
            Mobile-Former-52M & 3.5M & 0.05B & 68.7 \\
            Mobile-Former-96M & 4.6M & 0.1B & 72.8 \\
            Mobile-Former-214M & 9.4M & 0.2B & 76.7 \\
            \midrule
            Mobile-Former-508M & 14.0M & 0.5B & 79.3 \\
            \midrule \midrule
            \tinyname-0 & 3.4M  & 0.8B & 75.5 \\
            \tinyname-1 & 5.1M  & 1.2B & 78.3 \\
            \tinyname-2 & 9.8M & 2.3B & 81.0 \\
            \midrule
            \tinyname-3 & 19.5M & 4.5B & 82.7 \\
            \end{tabular}}
        \end{center}}
    \end{minipage}
    \begin{minipage}{0.49\linewidth}{
        \begin{center}
            \scalebox{0.84}{\begin{tabular}{l c c c}
            input size $256^{2}$ & params & FLOPs & \multicolumn{1}{c}{top-1 acc.} \\
            \toprule
            MobileViT-XS & 2.3M & 0.7B & 74.8 \\
            MobileViT-S & 5.6M & 2.0B & 78.4 \\
            \cmidrule(lr){1-4}
            MobileViTv2-1.0 & 4.9M & 1.8B & 78.1 \\
            MobileViTv2-1.5 & 10.6M & 4.0B & 80.4 \\
            MobileViTv2-2.0 & 18.5M & 7.5B & 81.2 \\
            \midrule \midrule
            \tinyname-1 & 5.1M & 1.6B & 79.2 \\
            \tinyname-2 & 9.8M & 3.0B & 81.7 \\
            \tinyname-3 & 19.5M & 6.0B & 83.3 \\
            \end{tabular}}
        \end{center}}
    \end{minipage}
    \label{tab:i1k_result_tiny}
\end{table}

\section{Ablation Studies on ImageNet}

At micro level, we perform ablation studies on the MOAT block design and downsampling layer in the following and the order of MBConv and Attention in MOAT block in~\secref{sec:ablation_micro_appendix}. At macro level, we perform ablation studies on the MOAT-based model and MOAT meta architecture in~\secref{sec:ablation_macro_appendix}.

\textbf{MOAT block design.}
In~\tabref{tab: moat_block_ablation}, we ablate the MOAT block design, which only affects the last two stages of MOAT, and we keep everything else the same (\eg, training recipes).
We start from the Transformer block, consisting of Attn (self-attention) and MLP, which already attains a strong top-1 accuracy (82.6\%).
Directly inserting a $3\times3$ depthwise convolution in the MLP degrades the performance by 0.9\%.
If we additionally insert batch normalization and GeLU between convolutions (\ie, replace MLP with MBConv, but no Squeeze-and-Excitation), the performance is improved to 82.9\%.
Finally, placing MBConv before Attn reaches the performance of 83.3\%.
Additionally, our MOAT block brings more improvements (from 1.2\% up to 2.6\% gains) in the tiny model regime.

\begin{table}[!t]
\small
    \centering
    \caption{\textbf{Ablation studies of MOAT block design} on ImageNet-1K with input size 224.
    }
    \label{tab: moat_block_ablation}

    \begin{minipage}{0.49\linewidth}{
        \begin{center}
        \scalebox{0.7}{
        \begin{tabular}{c l|ccc}
        model & block composition & params & FLOPs & top-1 acc.\\
        \toprule
        \multirow{4}{*}{MOAT-0} & Attn + MLP & 28.0M & 5.4B & 82.6\\
        & Attn + MLP (w/ depth. conv) & 28.2M & 5.4B & 81.7\\
        & Attn + MBConv  & 28.2M & 5.4B & 82.9\\
        & \baseline{MBConv + Attn} & \baseline{27.8M} & \baseline{5.7B} & \baseline{83.3}\\
        \end{tabular}}
        \end{center}}
    \end{minipage}
    \begin{minipage}{0.49\linewidth}{
        \begin{center}
        \scalebox{0.7}{
        \begin{tabular}{c l|ccc}
        model & block composition & params & FLOPs & top-1 acc.\\
        \toprule
        \multirow{2}{*}{tiny-MOAT-2} & Attn + MLP & 9.8M & 2.2B & 79.8 \\
        & \baseline{MBConv + Attn} & \baseline{9.8M} & \baseline{2.3B} & \baseline{81.0}\\
        \midrule
        \multirow{2}{*}{tiny-MOAT-1} & Attn + MLP & 5.1M & 1.1B & 76.2 \\
        & \baseline{MBConv + Attn} & \baseline{5.1M} & \baseline{1.2B} & \baseline{78.3}\\
        \midrule
        \multirow{2}{*}{tiny-MOAT-0} & Attn + MLP & 3.3M & 0.8B & 72.9 \\
        & \baseline{MBConv + Attn} & \baseline{3.4M} & \baseline{0.8B} & \baseline{75.5}\\
        \end{tabular}}
        \end{center}}
    \end{minipage}
\end{table}

\textbf{Downsampling layer.}
For the MOAT block design,
we do not need the extra downsampling layer like (1) average-pooling in CoAtNet~\citep{dai2021coatnet}, (2) patch-embedding layer (\ie, $2\times2$ convolution with stride 2) in Swin~\citep{liu2021swin} and ConvNeXt~\citep{liu2022convnet}, or
(3) strided depthwise convolution in PiT~\citep{heo2021rethinking} and RegionViT~\citep{chen2022regionvit}.
As shown in~\tabref{tab:moat_downsampling_ablation}, using patch-embedding layer indeed improves over the average-pooling scheme by 0.2\% accuracy, but it takes more cost of model parameters.
Additionally, using the strided depthwise convolution for downsampling leads to 0.2\% worse performance than the patch-embedding layer.
By contrast, our MOAT design (\ie, delegating the downsampling to the MBConv block) shows the best performance with the least cost of parameters and comparable FLOPs.

\begin{table}[!ht]
    \centering
    \caption{\textbf{Ablation studies of the downsampling layer design} on ImageNet-1K, using MOAT-0 and input size 224. We compare our MOAT design (in grey) with (1) CoAtNet (using average-pooling for downsampling), (2) Swin/ConvNeXt designs (using strided $2\times2$ convolution for downsampling), and (3) PiT/RegionViT designs (using strided $3\times3$ depthwise convolution for downsampling).
    }
    \label{tab:moat_downsampling_ablation}
    \scalebox{0.75}{
    \begin{tabular}{rc|ccc}
    block composition & downsampling type & params (M) & FLOPs (B) & top-1 acc.\\
    \toprule
    AveragePooling + Attn + MLP & CoAtNet & 28.0 & 5.4 & 82.6 \\
    PatchEmbedding + Attn + MLP & Swin, ConvNeXt & 30.2 & 5.6 & 82.8 \\
    StridedDepthConv + Attn + MLP & PiT, RegionVit & 28.8 & 5.5 & 82.6 \\
    \baseline MBConv + Attn & \baseline MOAT & \baseline 27.8 & \baseline 5.7 & \baseline 83.3 \\
    \end{tabular}
    }
    \label{tab: ablation_dowmsampling_layer}
\end{table}

\section{Related Work}

Transformers~\citep{vaswani2017attention} were recently introduced to the vision community~\citep{wang2018non,parmar2019stand,hu2019local} and demonstrated remarkable performance on vision recognition tasks~\citep{carion2020end,zhu2020deformable,wang2021max,arnab2021vivit,liu2021swin,cheng2021per,yu2022cmt,kim2022tubeformer,cheng2021masked,yu2022kmeans}, thanks to their ability to efficiently encode long-range interaction via the attention mechanism~\citep{bahdanau2014neural}. Particularly, ViT~\citep{dosovitskiy2020image} obtains impressive results on ImageNet~\citep{russakovsky2015imagenet} by applying the vanilla Transformer with the novel large stride patch embedding, after pretraining on the proprietary large-scale JFT dataset~\citep{sun2017revisiting}. There have been several works aiming to improve the vision transformers, either with better training strategies~\citep{touvron2021training,touvron2021going,steiner2021train,zhai2022scaling,touvron2022deit} or with efficient local-attention modules~\citep{huang2019ccnet,ho2019axial,wang2020axial,liu2021swin,chu2021twins,yang2021focal,yu2021glance,dong2022cswin,tu2022maxvit}.

Since the debut of AlexNet~\citep{krizhevsky2012imagenet}, the vision community has witnessed a rapid improvement on the ImageNet benchmark using different types of ConvNets, including (but not limited to) VGGNet~\citep{simonyan2015very}, Inceptions~\citep{szegedy2015going,ioffe2015batch,szegedy2016rethinking,szegedy2017inception}, ResNets~\citep{he2016deep,he2016identity}, ResNeXt~\citep{xie2017aggregated}, DenseNet~\citep{huang2017densely}, SENet~\citep{hu2018squeeze},
MobileNets~\citep{howard2017mobilenets,sandler2018mobilenetv2,howard2019searching}, EfficientNets~\citep{tan2019efficientnet,tan2021efficientnetv2}, and ConvNeXt~\citep{liu2022convnet} each focusing on different aspects of accuracy and efficiency. The ubiquity of ConvNets in computer vision could be attributed to their built-in inductive biases. %

Given the success of Transformers and ConvNets, another line of research is to explore how to effectively combine them. Swin~\citep{liu2021swin,liu2022swin}, PVT~\citep{wang2021pyramid,wang2022pvt},  MViT~\citep{fan2021multiscale,li2022improved}, and PiT~\citep{heo2021rethinking} adopt the ConvNet hierarchical structure to extract multi-scale features for Transformers. SASA~\citep{parmar2019stand}, AA-ResNet~\citep{bello2019attention}, Axial-ResNet~\citep{wang2020axial} and BoTNet~\citep{srinivas2021bottleneck} incorporate the attention modules to ResNets. CvT~\citep{wu2021cvt}, LeViT~\citep{graham2021levit}, Visformer~\citep{chen2021visformer}, and $\text{ViT}_{C}$~\citep{xiao2021early} replace ViT's patch embedding with strided convolutions.
CeiT~\citep{yuan2021incorporating} and CMT~\citep{guo2022cmt} incorporate depthwise convolution to the transformer block's MLP.
ViTAE~\citep{xu2021vitae} adopts parallel attention modules and convolutional layers. LVT~\citep{yang2022lite} introduces local self-attention into the convolution.
Recently, CoAtNet~\citep{dai2021coatnet} and MobileViT~\citep{mehta2021mobilevit} propose hybrid models that build on top of the efficient Mobile Convolution~\citep{sandler2018mobilenetv2} and Transformer block.

\paragraph{Acknowledgements}
We thank Wen-Sheng Chu for the support and discussion.

\bibliography{iclr2023_conference}
\bibliographystyle{iclr2023_conference}

\clearpage

\appendix
\section{Appendix}
In the appendix, we provide more details for both our model and experiments. 

\begin{itemize}
    \item In~\secref{sec:moat_implementation}, we provide MOAT implementation details.
    \item In~\secref{sec:imagenet_classification_appendix}, we provide ImageNet experimental details.
    \item In~\secref{sec:imagenet_v2}, we provide ImageNet-V2 experimental results.
    \item In~\secref{sec:coco_detection_details_appendix}, we provide COCO detection experimental details.
    \item In~\secref{sec:coco_detection_results_appendix}, we provide more COCO object detection experimental results.
    \item In~\secref{sec:ade20k_semantic_appendix}, we provide ADE20K semantic segmentation experimental detaills.
    \item In~\secref{sec:coco_panoptic_appendix}, we provide COCO panoptic segmentation experiments.
    \item In~\secref{sec:ablation_micro_appendix}, we provide ablation studies on the MOAT micro-level design.
    \item In~\secref{sec:ablation_macro_appendix}, we provide ablation studies on the MOAT macro-level design.
    \item In~\secref{sec:measurements}, we provide the ImageNet trainng time, peak training memory and throughput measurement of MOAT models.
    \item In~\secref{sec:limitations_appendix}, we discuss limitations of our model.
\end{itemize}

\subsection{MOAT implementation details}
\label{sec:moat_implementation}
In the MOTA networks, we employ kernel size 3 for both convolutions and depthwise convolutions. We use the multi-head self attention~\citep{vaswani2017attention}, where each attention head has channels 32.
For the MBConv and MOAT blocks, we use expansion ratio 4.
The SE module~\citep{hu2018squeeze} in the MBConv blocks (\ie, 2nd and 3rd stages) adopt reduction ratio 4 (relative to the input channels). 

Our MOAT block includes the relative positional embedding~\citep{shaw2018self,dai2021coatnet} for ImageNet.
However, the downstream tasks usually take a larger input resolution than ImageNet, demanding for a special adaptation (\eg, bilinear interpolation of pretrained positional embedding).
For simplicity, we remove the positional embedding, when running MOAT on downstream tasks.

\subsection{ImageNet Image Classification}

\subsubsection{ImageNet Experiments}
\label{sec:imagenet_classification_appendix}

The ImageNet-1K dataset~\citep{russakovsky2015imagenet} contains 1.2M training images with 1000 classes.
We report top-1 accuracy on the ImageNet-1K validation set, using the last checkpoint.
We also experiment with pretraining on the larger ImageNet-22K dataset, and then fine-tuning on the ImageNet-1K.
We closely follow the prior works~\citep{dai2021coatnet,liu2022convnet} and provide more details below.
In~\tabref{tab:i1k21k_result_full}, we compare our MOAT with more state-of-the-art models.

\begin{table}[!ht]
\small
    \centering
    \caption{\textbf{Performance on ImageNet-1K} with more state-of-the-art models are included. \textbf{1K only:} Using ImageNet-1K only. \textbf{22K + 1K:} ImageNet-22K pretraining and ImageNet-1K fine-tuning. 
    }
    \scalebox{0.85}{\begin{tabular}{c l c c c c c}
        \multicolumn{2}{c}{model} & eval size & params & FLOPs &  \multicolumn{2}{c}{ImageNet-1K top-1 accuracy} \\
        \toprule
        \multicolumn{5}{c}{} & 1K only & 22K+1K \vspace{0.5em} \\
        \multirow{10}{*}{ConvNets}
        & RegNetY-16G~\citep{radosavovic2020designing} & $224^2$ & 84M & 16.0B & 82.9 & - \\
        & NFNet-F5~\citep{brock2021high} & $544^2$ & 377M & 289.8B & 86.0 & - \\
        \cmidrule(lr){2-7}
        & EfficientNetV2-S~\citep{tan2021efficientnetv2} & $480^2$ & 22M & 8.8B & 83.9 & 84.9 \\
        & EfficientNetV2-M~\citep{tan2021efficientnetv2} & $480^2$ & 54M & 24B & 85.1 & 86.2 \\
        & EfficientNetV2-L~\citep{tan2021efficientnetv2} & $480^2$ & 120M & 53B & 85.7 & - \\
        & EfficientNetV2-XL~\citep{tan2021efficientnetv2} & $480^2$ & 208M & 94B & - & 87.3 \\
        \cmidrule(lr){2-7}
        & ConvNeXt-T~\citep{liu2022convnet} & $224^2$ & 29M & 4.5B & 82.1 & 82.9 \\
        & ConvNeXt-S~\citep{liu2022convnet} & $224^2$ & 50M & 8.7B & 83.1 & 84.6 \\
        & ConvNeXt-B~\citep{liu2022convnet} & $224^2$ & 89M & 15.4B & 83.8 & 85.8 \\
        & ConvNeXt-L~\citep{liu2022convnet} & $384^2$ & 198M & 101.0B & 85.5 & 87.5 \\
        & ConvNeXt-XL~\citep{liu2022convnet} & $384^2$ & 350M & 179.0B & 85.5 & 87.8 \\
        \cmidrule(lr){1-7}
        \multirow{12}{*}{ViTs}
        & DeiT-B~\citep{touvron2021training} & $384^2$ & 86M  & 55.4B & 83.1 & - \\
        & CaiT-S-36~\citep{touvron2021going} & $384^2$ & 68M & 48.0B & 85.0 & - \\
        & DeepViT-L~\citep{zhou2021deepvit} & $224^2$ & 55M & 12.5B &  83.1 & - \\
        & PVT-Large~\citep{wang2021pyramid}  & $224^2$ & 61.4M  & 9.8B & 81.7 & - \\
        & HaloNet-H4~\citep{vaswani2021scaling}  & $384^2$ & 85M  & - & 85.6 & - \\
        & HaloNet-H5~\citep{vaswani2021scaling}  & $512^2$ & 85M  & - & 85.8 & - \\
        \cmidrule(lr){2-7}
        & Swin-T~\citep{liu2021swin}  & $224^2$ & 28M  & 4.5B & 81.3  & - \\
        & Swin-S~\citep{liu2021swin}  & $224^2$ & 50M  & 8.7B & 83.0  & - \\
        & Swin-B~\citep{liu2021swin}  & $224^2$ & 88M  & 15.4B & 83.5  & 85.2 \\
        & Swin-L~\citep{liu2021swin}  & $384^2$ & 197M & 103.9B & -  & 87.3 \\
        & SwinV2-L~\citep{liu2022swin}  & $384^2$ & 197M & 115.4B & - & 87.7 \\
        \cmidrule(lr){2-7}
        & Focal-B~\citep{yang2021focal}  & $224^2$ & 89.8M & 16.0B & 83.8  & - \\
        & CSwin-B~\citep{dong2022cswin}  & $384^2$ & 78M & 47.0B & 85.4  & 87.0 \\
        & CSwin-L~\citep{dong2022cswin}  & $384^2$ & 173M & 96.8B & -  & 87.5 \\
        & MViTv2-H~\citep{li2022improved}  & $512^2$ & 667M & 763.5B & - & 88.8 \\
        \cmidrule(lr){1-7}
        \multirow{11}{*}{Hybrid} 
        & BotNet-T7~\citep{srinivas2021bottleneck}  & $384^2$ & 75.1M & 45.8B & 84.7 & - \\
        & LambdaResNet-420~\citep{bello2021lambdanetworks} & $320^2$ & - & - & 84.9 & - \\
        & T2T-ViT-24~\citep{yuan2021tokens} & $224^2$ & 64.1M & 15.0B & 82.6 & - \\
        & CMT-S~\citep{guo2022cmt} & $224^2$ & 25.1M & 4.0B & 83.5 & - \\
        & CeiT-S~\citep{yuan2021incorporating}   & $384^2$ & 24.2M & 12.9B & 83.3 & - \\
        & CvT-21~\citep{wu2021cvt}   & $384^2$ & 32M & 24.9B & 83.3 & - \\
        & PVTv2-B5~\citep{wang2022pvt} & $224^2$ & 82M &  11.8B & 83.8 & - \\
        & MaxViT-XL~\citep{tu2022maxvit}  & $512^2$ & 475M & 535.2B & - & 88.7 \\
        \cmidrule(lr){2-7}
        & CoAtNet-0~\citep{dai2021coatnet} & $224^2$ & 25M  & 4.2B & 81.6 & - \\
        & CoAtNet-1~\citep{dai2021coatnet} & $224^2$ & 42M  & 8.4B & 83.3 & -  \\
        & CoAtNet-2~\citep{dai2021coatnet} & $224^2$ & 75M  & 15.7B & 84.1 & - \\
        & CoAtNet-3~\citep{dai2021coatnet} & $384^2$ & 168M & 107.4B & 85.8 & 87.6\\
        & CoAtNet-4~\citep{dai2021coatnet} & $512^2$ & 275M & 360.9B & - & 88.6 \\
        \midrule \midrule
        \multirow{12}{*}{\makecell{\bf Hybrid \\ \bf (ours)}} 
        & \name-0 & $224^2$ & 27.8M  & 5.7B & 83.3 & 83.6 \\
        & \name-1 & $224^2$ & 41.6M  & 9.1B & 84.2 & 84.9  \\
        & \name-2 & $224^2$ & 73.4M & 17.2B & 84.7 & 86.0 \\
        & \name-3 & $224^2$ & 190.0M & 44.9B & 85.3 & 86.8 \\
        \cmidrule(lr){2-7}%
        & \name-0 & $384^2$ & 27.8M & 18.2B & 84.6  & 85.7 \\
        & \name-1 & $384^2$ & 41.6M  & 29.6B & 85.9  & 87.0 \\
        & \name-2 & $384^2$ & 73.4M  & 54.3B & 86.2 & 87.5 \\
        & \name-3 & $384^2$ & 190.0M & 141.2B & 86.5 & 88.2 \\
        \cmidrule(lr){2-7}%
        & \name-1 & $512^2$ & 41.6M & 58.7B & 86.2 & 87.2 \\
        & \name-2 & $512^2$ & 73.4M & 104.6B  & 86.5 & 87.7 \\
        & \name-3 & $512^2$ & 190.0M & 271.0B & 86.7 &  88.4 \\
        \cmidrule(lr){2-7}%
        & \name-4 & $512^2$ & 483.2M & 648.5B & - &  89.1 \\
    \end{tabular}}
    \label{tab:i1k21k_result_full}
\end{table}

\begin{table}[!ht]
    \centering
    \caption{\textbf{Performance on ImageNet-1K-V2}.
    }
    \label{tab:imagevet_1k_v2}
    \scalebox{0.8}{
    \begin{tabular}{cccccc}
    {model} & params & input size & FLOPs &  \multicolumn{2}{c}{ImageNet-1K-V2 top-1 accuracy (\%)} \\
    \toprule
    \multicolumn{4}{c}{} & 1K only & 22K + 1K \vspace{0.5em} \\
    LeViT-256~\citep{graham2021levit} & 256 & 18.9M & 1.1B & 70.0 & -- \\
    DeiT-B~\citep{touvron2021training} & 224 & 86M & 17.5B & 71.5 & -- \\
    CaiT-S36~\citep{touvron2021going} & 224 & 68M & 13.9B & 72.5 & -- \\
    SwinV2-B~\citep{liu2021swin} & 384 & 88M & 54.7B & -- & 78.1 \\
    SwinV2-L~\citep{liu2021swin} & 384 & 197M & 115.4B & -- & 78.3 \\
    \midrule
    \midrule
    \tinyname-0 & 224 & 3.4M & 0.8B & 64.3 & -- \\
    \tinyname-1 & 224 & 5.1M & 1.2B & 67.3 & -- \\
    \tinyname-2 & 224 & 9.8M  & 2.3B  & 70.1 & -- \\
    \tinyname-3 & 224 & 19.5M & 4.5B  & 72.1 & -- \\
    \cmidrule(lr){1-6}
    \tinyname-1 & 256 & 5.1M & 1.6B & 68.2 & -- \\
    \tinyname-2 & 256 & 9.8M & 3.0B & 70.9 & -- \\
    \tinyname-3 & 256 & 19.5M & 6.0B & 72.9 & -- \\
    \cmidrule(lr){1-6}
    \name-0 & 224 & 27.8M & 5.7B & 72.8 & 74.1 \\
    \name-1 & 224 & 41.6M & 9.1B & 74.2 & 75.8 \\
    \name-2 & 224 & 73.4M & 17.2B & 74.3 & 76.7 \\
    \name-3 & 224 & 190.0M & 44.9B & 75.5 & 78.4 \\
    \cmidrule(lr){1-6}
    \name-0 & 384 & 27.8M & 18.2B & 74.5 & 76.4 \\
    \name-1 & 384 & 41.6M & 29.6B & 76.2 & 78.1 \\
    \name-2 & 384 & 73.4M & 54.3B & 76.5 & 78.7 \\
    \name-3 & 384 & 190.0M & 141.2B & 77.5 & 80.0 \\
    \cmidrule(lr){1-6}
    \name-1 & 512 & 41.6M & 58.7B & 76.8 & 78.4 \\
    \name-2 & 512 & 73.4M & 104.6B & 77.1 & 79.3 \\
    \name-3 & 512 & 190.0M & 271.0B & 77.8 & 80.6 \\
    \cmidrule(lr){1-6}
    \name-4 & 512 & 483.2M & 648.5B & -- & 81.5 \\
    \end{tabular}}
\end{table}

\textbf{Experimental setup.}
We train \name models on ImageNet-1K with resolution 224 for 300 epochs.
If pretraining on the larger ImageNet-22K, we use resolution 224 and 90 epochs.
Afterwards, the models are fine-tuned on ImageNet-1K for 30 epochs.
During fine-tuning, we also experiment with larger resolutions (\eg, 384 and 512).
We employ the typical regularization methods during training, such as label smoothing~\citep{szegedy2016rethinking}, RandAugment~\citep{cubuk2020randaugment}, MixUp~\citep{zhang2017mixup}, stochastic depth~\citep{huang2016deep}, and Adam~\citep{kingma2014adam} with decoupled weight decay (\ie, AdamW~\citep{loshchilov2019decoupled}).
See~\tabref{tab:main_hparam} and~\tabref{tab:tiny_moat_hparam} for detailed hyper-parameters.

\begin{table}[!ht]
\caption{\textbf{ \name ImageNet hyper-parameter settings.} %
}
\centering
\scalebox{0.85}{
\begin{tabular}{l|cc|cc}
    \multirow{4}{*}{\bf hyper-parameter} 
    & \multicolumn{2}{c|}{\bf ImageNet-1K} 
    & \multicolumn{2}{c}{\bf ImageNet-22K} \\
    & 1K & 1K $\rightarrow$ 1K & 22K & 22K $\rightarrow$ 1K \\
    & pre-training & fine-tuning & pre-training & fine-tuning \\
    & \multicolumn{2}{c|}{(\name-0/1/2/3)} 
    & \multicolumn{2}{c}{(\name-0/1/2/3)} \\
    \toprule
    \multirow{2}{*}{stochastic depth rate} 
    & 0.2 / 0.3 & 0.2 / 0.3 & 0.1 / 0.2 & 0.1 / 0.2 \\
    {} & / 0.5 / 0.7 & / 0.5 / 0.9 & / 0.3 / 0.6 & / 0.3 / 0.6\\
    \midrule
    center crop           & true      & false   & true     & false \\
    randaugment           & 2, 15     & 2, 15/15/15/20   &
    2, 5 & 2, 5 \\
    mixup alpha           & 0.8       & 0.8     & none     & none     \\
    loss type             & softmax   & softmax & sigmoid  & softmax  \\
    label smoothing       & 0.1       & 0.1     & 0.0001   & 0.1      \\
    train epochs          & 300       & 30      & %
    90 & 30       \\
    train batch size      & 4096      & 512     & 4096     & %
    1024     \\
    optimizer type        & AdamW     & AdamW   & AdamW    & AdamW    \\
    peak learning rate    & 3e-3      & 5e-5    & 1e-3     & %
    5e-5     \\
    min learning rate     & 1e-5      & 5e-5    & 1e-5     & %
    5e-5    \\
    warm-up               & 10K steps & none    & 5 epochs & none        \\
    lr decay schedule     & cosine    & none    & linear   & none    \\
    weight decay rate     & 0.05      & 1e-8    & 0.01     & 1e-8     \\
    gradient clip         & 1.0       & 1.0     & 1.0      & 1.0      \\
    EMA decay rate        & 0.9999    & 0.9999  & None     & 0.9999   \\
\end{tabular}}
\label{tab:main_hparam}
\end{table}

\begin{table}[!ht]
\caption{\textbf{\tinyname ImageNet hyper-parameter settings.} $^{\star}$: use EMA decay rate 0.9999 for \tinyname-3.}
\centering
\scalebox{0.85}{
\begin{tabular}{l|cc}
    \multirow{4}{*}{\bf hyper-parameter} 
    & \multicolumn{2}{c}{\bf ImageNet-1K} \\
    & 1K & 1K  \\
    & input size 224 & input size 256  \\
    & \multicolumn{2}{c}{(\tinyname-0/1/2/3)} \\
    \toprule
    \multirow{2}{*}{stochastic depth rate} 
    & 0.0 / 0.0 & 0.0 / 0.0 \\
    {} & / 0.0 / 0.1 & / 0.0 / 0.1 \\
    \midrule
    center crop           & true      & true    \\
    randaugment           & 2, 15     & 2, 15   \\
    mixup alpha           & 0.8       & 0.8     \\
    loss type             & softmax   & softmax \\
    label smoothing       & 0.1       & 0.1     \\
    train epochs          & 300       & 300     \\
    train batch size      & 4096      & 4096    \\
    optimizer type        & AdamW     & AdamW   \\
    peak learning rate    & 3e-3      & 3e-3    \\
    min learning rate     & 1e-5      & 1e-5    \\
    warm-up               & 10K steps & 10K steps   \\
    lr decay schedule     & cosine    & cosine      \\
    weight decay rate     & 0.05      & 0.05        \\
    gradient clip         & 1.0       & 1.0         \\
    EMA decay rate        & None$^{\star}$      & None$^{\star}$        \\
\end{tabular}}
\label{tab:tiny_moat_hparam}
\end{table}

\subsubsection{ImageNet-1K-V2 Evaluation}
\label{sec:imagenet_v2}

To further demonstrate the transferability and generalizability of our MOAT models, we perform additional evaluations on the ImageNet-1K-V2~\citep{recht2019imagenet}, using our ImageNet~\citep{russakovsky2015imagenet} pretrained checkpoints.
We report an extensive evaluation, using MOAT and several input resolutions, on ImageNet-1K-V2, aiming to establish another solid baseline for the community, as we notice that most of the existing models do not report results on ImageNet-1K-V2.
As shown in the~\tabref{tab:imagevet_1k_v2}, MOAT does not overfit to ImageNet-1K-V1 dataset and generalizes well to ImageNet-1K-V2 dataset, as we observe a continuous performance improvement from small to large models.
Under the fair comparison, with ImageNet-22K pretrainng and input size 384, MOAT-2/3 surpass the current state-of-the-art model SwinV2-B/L by 0.6/1.7\%, respectively.
Additionally, our MOAT-4, with input size 512, achieves a new state-of-the-art performance of 81.5\%, without extra proprietary training data.

\subsection{COCO Object Detection and Instance Segmentation}
\label{sec:coco_appendix}

\subsubsection{COCO Object Detection Experimental Details}
\label{sec:coco_detection_details_appendix}
\textbf{Experimental setup.}
We train Cascade Mask R-CNN~\citep{cai2018cascade,he2017mask} on the COCO 2017 dataset~\citep{lin2014microsoft} with our \name architectures.
The dataset contains 118K training and 5K validation samples.
We use the official TensorFlow~\citep{tensorflow-osdi2016} implementation of Cascade Mask R-CNN by TF-Vision Model Garden~\citep{tensorflowmodelgarden2020}.
Our training setting closely follows the prior works~\citep{chen2022simple,tu2022maxvit}, except that we use batch size 64 and initial learning rate 0.0001.
To adapt the \name models to high-resolution inputs, we partition the features into non-overlapping windows for the self-attention computations with the window size set to 14 for the second last stage, and use global attention for the last stage.
As a result of this window partition, the input size must be divisible by 14.
The TF-Vision Model Garden codebase further requires the input size to be square (with padding) and divisible by 64.
Hence, we choose 1344 as the input size, similar to the size used in the baseline methods (\ie, longest side is no more than 1333).
We use Feature Pyramid Network~\citep{lin2017feature} to integrate features from different levels.

\subsubsection{More COCO Object Detection Experimental Results}
\label{sec:coco_detection_results_appendix}

In this section, we perform more COCO object detection experiments with 896 input size.
All the backbone are pretrained on ImageNet-1K dataset. MOAT-0/1/2 surpass UViT~\citep{chen2021simple} and MaxViT~\citep{tu2022maxvit} by 3.9/5.2/4.9\% AP$^{\text{box}}$ (3.1/4.1/3.9\% AP$^{\text{mask}}$), and 3.0/4.0/4.0\% AP$^{\text{box}}$ (2.4/3.2/3.0\% AP$^{\text{mask}}$), respectively.

\begin{table}[!ht]
    \centering
    \caption{\textbf{Object detection and instance segmentation on the COCO 2017 \texttt{val} set}.
    We employ Cascade Mask-RCNN, and single-scale inference (hard NMS). All backbones are pretrained on ImageNet-1K.
    }
    \label{tab: more_coco_exp}
    \scalebox{0.76}{
    \begin{tabular}{l|ccc|ccc|ccc}
    backbone & input size & params & FLOPs & AP$^{\text{box}}$ & AP$^{\text{box}}_{\text{50}}$ & AP$^{\text{box}}_{\text{75}}$ & AP$^{\text{mask}}$ & AP$^{\text{mask}}_{\text{50}}$ & AP$^{\text{mask}}_{\text{75}}$ \\
    \toprule
    UViT-T~\citep{chen2021simple} & $896\times896$ & 51M & 720B & 51.2 & -- & -- & 43.9 & -- & -- \\
    UViT-S~\citep{chen2021simple} & $896\times896$ & 59M & 882B & 51.9 & -- & -- & 44.5 & -- & -- \\
    UViT-B~\citep{chen2021simple} & $896\times896$ & 74M & 1160B & 52.5 & -- & -- & 44.8 & -- & -- \\
    \midrule
    MaxViT-T~\citep{tu2022maxvit} & $896\times896$ & 86M & 475B & 52.1 & 71.9 & 56.8 & 44.6 & 69.1 & 48.4 \\
    MaxViT-S~\citep{tu2022maxvit} & $896\times896$ & 108M & 595B & 53.1 & 72.5 & 58.1 & 45.4 & 69.8 & 49.5 \\
    MaxViT-B~\citep{tu2022maxvit} & $896\times896$ & 146M & 856B & 53.4 & 72.9 & 58.1 & 45.7 & 70.3 & 50.0 \\
    \midrule \midrule
    \name-0 & $896\times896$ & 65M & 525B & 55.1 & 73.6 & 59.9 & 47.0 & 70.5 & 51.1 \\
    \name-1 & $896\times896$ & 79M & 580B & 57.1 & 75.7 & 62.6 & 48.6 & 72.9 & 52.7 \\
    \name-2 & $896\times896$ & 110M & 710B & 57.4 & 76.0 & 63.0 & 48.7 & 73.2 & 53.1 \\
    \end{tabular}
    }
\end{table}

\subsection{ADE20K Semantic Segmentation}
\label{sec:ade20k_semantic_appendix}
\textbf{Experimental setup.}
We experiment with the proposed \name models on ADE20K semantic segmentation dataset~\citep{zhou2019semantic} using DeepLabv3+~\citep{deeplabv3plus2018,chen2018deeplabv2}. We fine-tune the global attention for \name. The same training strategies are used for all backbone variants. Specifically, for training hyper-parameters, we train the model with 32 TPU cores for 180k iterations, with batch size 64, Adam~\citep{kingma2014adam} optimizer, and a poly schedule learning rate starting at 0.0001. For data augmentations, the inputs images are resized and padded to either $513\times 513$ or $641\times 641$, with random cropping, flipping, and color jittering~\citep{cubuk2018autoaugment}. \textbf{No test-time augmentation is used during inference.}

\subsection{COCO Panoptic Segmentation}
\label{sec:coco_panoptic_appendix}
\textbf{Experimental setup.}
We also evaluate the proposed \name architectures on the challenging COCO panoptic segmentation dataset~\citep{lin2014microsoft} using Panoptic-DeepLab~\citep{cheng2019panoptic} with the official codebase~\citep{deeplab2_2021}. We fine-tune the global attention on downstream segmentation tasks for \name. We adopt the same training strategies for \name and its counterparts. Specifically, for training hyper-parameters, we train the model with 32 TPU cores for 200k iterations with the first 2k for warm-up stage. We use batch size 64, Adam~\citep{kingma2014adam} optimizer, and a poly schedule learning rate starting at 0.0005. For data augmentations, the inputs images are resized and padded to $641\times 641$, with random cropping, flipping, and color jittering~\citep{cubuk2018autoaugment}. \textbf{No test-time augmentation is used during inference.}

\textbf{Main results.}
The results are summarized in \tabref{tab:coco_pan_exp}, where \name consistently outperforms other backbones. Specifically, our \name-0 surpasses ConvNeXt-T significantly by 4.3\% PQ. In the large model regime, \name-3 surpasses ConvNeXt-L by 3.5\%. Our \name-4 achieves the performance of 46.7\% PQ, outperforming the heavy backbone SWideRNet~\citep{swidernet_2020} by 2.3\%.

\begin{table}[!ht]
    \centering
    \caption{\textbf{Panoptic segmentation on COCO \texttt{val} set.} The results are obtained by applying different backbones with Panoptic-DeepLab, using \textbf{single-scale} inference (\ie, no test-time augmentation). Results for MobileNet, ResNet, and Xception are cited from~\citep{cheng2019panoptic}, and results for SWideRNet is cited from~\citep{swidernet_2020}, while results for ConvNeXt and \name are obtained using the official code-base~\citep{deeplab2_2021} with the same training recipe. All models are trained and evaluated with input images resized to $641\times641$, and thus FLOPs are also measured \wrt size $641\times 641$. \dag: use ImageNet-22K pretrained weights.
    }
    \label{tab:coco_pan_exp}
    \scalebox{0.8}{
    \begin{tabular}{l|cc|ccc}
    backbone & params & FLOPs & PQ (\%) & PQ\textsuperscript{Th} (\%)  & PQ\textsuperscript{St} (\%)   \\
    \toprule
    MobileNet-V3~\citep{howard2019searching} & - & 12.2B & 30.0 & - & - \\
    ResNet50~\citep{he2016deep} & - & 77.8B & 35.1 & - & - \\
    Xception-71~\citep{chollet2016xception} & - & 109.2B & 38.9 & - & - \\
    \midrule
    ConvNeXt-T~\citep{liu2022convnet} & 40.3M & 51.3B & 36.7 & 37.3 & 35.7 \\
    ConvNeXt-S~\citep{liu2022convnet} & 61.9M & 87.2B & 40.0 & 41.4 & 37.9  \\
    ConvNeXt-B\dag~\citep{liu2022convnet} & 103.8M & 146.2B & 41.7 & 43.6 & 38.9  \\
    ConvNeXt-L\dag~\citep{liu2022convnet} & 220.1M & 312.8B & 41.9 & 43.6 & 39.4  \\
    ConvNeXt-XL\dag~\citep{liu2022convnet} & 379.6M & 544.1B & 43.0 & 44.9 & 40.0  \\
    \midrule
    SWideRNet~\citep{swidernet_2020} & 752.5M & 2614.0B & 44.4 & - & - \\
    \midrule \midrule
    \name-0  & 39.5M & 76.8B & 41.0 & 42.6 & 38.6  \\
    \name-1  & 53.1M & 119.7B & 43.0 & 44.7 & 40.4 \\
    \name-2\dag  & 88.5M & 199.7B & 43.9 & 45.9 & 40.8 \\
    \name-3\dag  & 208.3M & 493.3B & 45.4 & 48.3 & 41.1 \\
    \name-4\dag  & 512.0M & 1134.7B & 46.7 & 49.5 & 42.4 \\
    \end{tabular}}
\end{table}

\subsection{More Ablation Studies}
\label{sec:more_ablation_studies}

\subsubsection{Ablation Studies on The MOAT Micro-level Design}
\label{sec:ablation_micro_appendix}
\textbf{Order of MBConv and Attn in MOAT block.}
Our MOAT block design reverses the order of Attention (Attn) and Mobile Convolution (MBConv), delegating the downsampling duty to the strided depthwise convolution within the MBConv. %
However, the dowsampling can be still performed in the MBConv with the \textit{original} order (\ie, Attn + MBConv).
Since the operations, Attn and MBConv, are interlaced, the key difference then comes from the first block in each stage, where the Attn is operated on the (1) spatially downsampled and/or (2) channel expanded features. %
To conduct the study, we employ different blocks in the MOAT  variants, using "Attn + MLP", "Attn + MBConv", or "MBConv + Attn". For the "Attn + MBConv" block, we further ablate the place (Attn \vs MBConv), where we apply the spatial downsampling and channel expansion operations.

In~\tabref{tab:mbconv_attn_order}, we observe the following results. First, replacing the MLP with MBConv improves the performance by 0.3\% and 0.7\% for MOAT-0 and tiny-MOAT-2. Second, if we perform both spatial downsampling and channel expansion at the MBConv block, the performance is further improved by 0.5\% and 0.9\% for MOAT-0 and tiny-MOAT-2, showing that MBConv learns better downsampled features. However, this design is equivalent to shifting the first Attn layer to its previous stage, reducing the representation capacity of the current stage. More concretely, only the last stage will be affected, since one layer is shifted. Third, to enhance the representation capacity, reversing the order of Attn and MBConv allows us to keep the first Attn layer in the same stage. This design further improves the performance by 0.7\% and 1.2\% for MOAT-0 and tiny-MOAT-2. Fourth, to compensate for the shifting effect, we could also employ another $1\times1$ convolution to expand the channels at the first Attn layer (then, MBConv only performs the spatial downsampling).
However, this design performs similarly to our MOAT block design, but uses more parameters and FLOPs.

\begin{table}[!t]
    \centering
    \caption{\textbf{Ablation studies of the order of MBConv and Attention (Attn)} on ImageNet-1K with input 224. We also ablate the place, where we apply the spatial downsampling and channel expansion.
    }
    \label{tab:mbconv_attn_order}
    \scalebox{0.65}{
    \begin{tabular}{cccc|ccc}
    model & block composition & spatial downsampling & channel expansion & params (M) & FLOPs (B) & top-1 acc.\\
    \toprule
    \multirow{5}{*}{MOAT-0} & Attn + MLP & Attn & Attn & 28.0 & 5.4 & 82.6\\
    & Attn + MBConv & Attn & Attn & 28.2 & 5.4 & 82.9\\
    & Attn + MBConv & MBConv & MBConv & 25.6 & 5.8 & 83.1\\
    & \baseline{MBConv + Attn} & \baseline{MBConv} & \baseline{MBConv} & \baseline{27.8} & \baseline{5.7} & \baseline{83.3}\\
    & Attn + MBConv & MBConv & Attn & 29.3 & 7.1 & 83.2\\
    \midrule
    \multirow{5}{*}{tiny-MOAT-2} & Attn + MLP & Attn & Attn & 9.8 & 2.2 & 79.8\\
    & Attn + MBConv & Attn & Attn & 9.9 & 2.2 & 80.5\\
    & Attn + MBConv & MBConv & MBConv & 9.0 & 2.3 & 80.7\\
    & \baseline{MBConv + Attn} & \baseline{MBConv} & \baseline{MBConv} & \baseline{9.8} & \baseline{2.3} & \baseline{81.0}\\
    & Attn + MBConv & MBConv & Attn & 10.3 & 2.8 & 81.0\\
    \end{tabular}}
\end{table}

\subsubsection{Ablation Studies on The MOAT Macro-level Design}
\label{sec:ablation_macro_appendix}
\textbf{Ablation studies on MOAT-based model.}
In~\tabref{tab: moat_moat_based_ablation}, we ablate the stage-wise design by using either MBConv or MOAT block in stage 2 to stage 5. The first stage is the convolutional stem, containing two $3\times3$ convolutions. We use the layer layout of MOAT-0. As shown in the table, the pure MOAT-based model (\ie, using MOAT blocks for all four stages) achieves the best performance of 83.6\%, which however uses the most FLOPs. Our MOAT model design (\ie, use MOAT block in the last two stages) attains the better trade-off between accuracy and model complexity.

\begin{table}[!ht]
    \centering
    \caption{\textbf{Ablation studies of MOAT-based model} on ImageNet-1K, using MOAT-0 layer layout and input size 224. We change the block type (MBConv \vs MOAT block) from stage 2 to stage 5. The first stage is fixed to use the convolutional stem.
    }
    \label{tab: moat_moat_based_ablation}
    \scalebox{0.86}{
    \begin{tabular}{ccccccc}
    stage-2 & stage-3 & stage-4 & stage-5 & params (M) & FLOPs (B) & top-1 acc.\\
    \toprule
    MOAT & MOAT & MOAT & MOAT & 28.2 & 11.9 & 83.6 \\
    MBConv & MOAT & MOAT & MOAT & 28.1 & 6.9 & 83.5 \\
    \baseline{MBConv} & \baseline{MBConv} & \baseline{MOAT} & \baseline{MOAT} & \baseline{27.8} & \baseline{5.7} & \baseline{83.3} \\
    MBConv & MBConv & MBConv & MOAT & 25.7 & 4.7 & 82.2 \\
    MBConv & MBConv & MBConv & MBConv & 23.4 & 4.5 & 82.0 \\
    \end{tabular}}
\end{table}

\begin{figure}[!ht]
  \centering
  \includegraphics[width=0.8\textwidth]{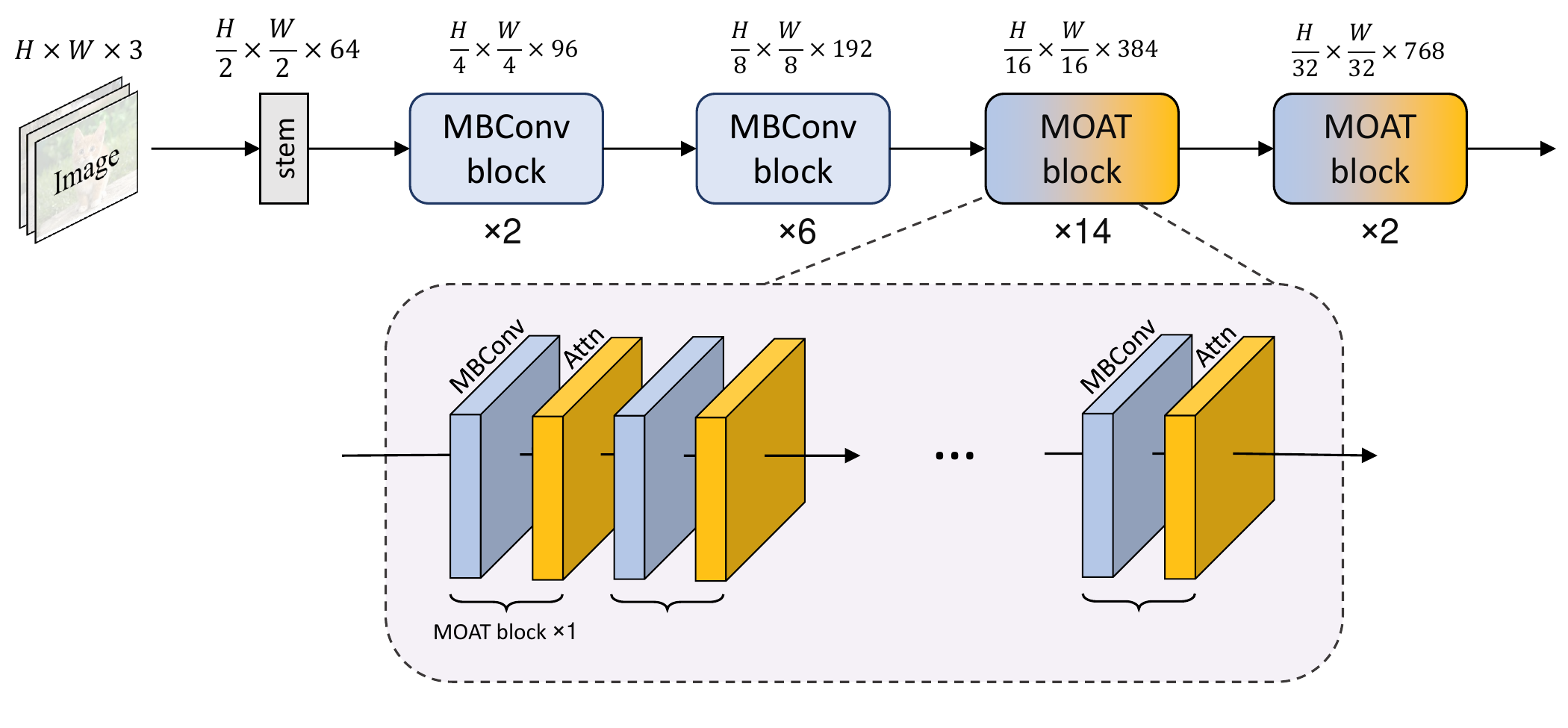}
  \caption{\textbf{Architecture of MOAT-1}, including the convolutional stem, MBConv, and MOAT blocks. %
  }
  \label{fig:meta_moat1}
\end{figure}

\textbf{Ablation studies on MOAT meta architecture.}
We perform ablation studies on the meta-architecture by varying the number of blocks per stage.
For simplicity, we only vary the block numbers in the third and fourth stages, while keeping the block numbers in the other stages unchanged.
Note that the first stage corresponds to the convolutional stem.
The studies with MOAT-1 meta architecture are shown in~\tabref{tab: moat_meta_architecture_ablation}.
In the end, we choose the layout \{2, 2, 6, 14, 2\} because it has the best performance and lower parameter cost.
Interestingly, our discovery echoes the layer layout proposed by CoAtNet~\citep{dai2021coatnet}.
We visualize the architecture of MOAT-1 in~\figureref{fig:meta_moat1}.

\begin{table}[!ht]
    \centering
    \caption{\textbf{Ablation studies of MOAT meta-architecture design} on ImageNet-1K, using MOAT-1 and input size 224. We control the first, second and last stages to have two blocks, and vary the block numbers of the third and fourth stages.
    }
    \label{tab: moat_meta_architecture_ablation}
    \scalebox{0.8}{
    \begin{tabular}{c|ccc}
    number of blocks in five stages & params (M) & FLOPs (B) & top-1 acc.\\
    \toprule
    (2, 2, 2, 16, 2) & 43.7 & 8.9 & 84.1 \\
    (2, 2, 4, 15, 2) & 42.6 & 9.0 & 84.2 \\
    \baseline (2, 2, 6, 14, 2) & \baseline 41.6 & \baseline 9.1 & \baseline 84.2 \\
    (2, 2, 8, 13, 2) & 40.6 & 9.2 & 84.1 \\
    (2, 2, 10, 12, 2) & 39.5 & 9.3 & 84.1 \\
    \end{tabular}}
\end{table}

\subsection{ImageNet Training Time, Peak Training Memory and Troughput Measurements}
\label{sec:measurements}

\begin{table}[!ht]
    \centering
    \caption{\textbf{ImageNet training time} measured in hours. We use 16 TPUv4 cores for training \name-\{0,1,2\} and 32 TPUv4 cores for \name-3. \name is training efficient:
    for ImageNet-22k pretraining, \name takes no more than 2.05 days, while for ImageNet-1k pretraining, \name takes $<$ 1 day. 
    }
    \label{tab: tpu_core_days}
    \scalebox{0.8}{
    \begin{tabular}{c|c|c|cc}
    dataset & model & pre-training 
    & \multicolumn{2}{c}{fine-tuning} \\
    \toprule
    \multirow{5}{*}{ImageNet-1K} & input size & $224 \times 224$ & $224 \times 224$ & $384 \times 384$  \\
    \cmidrule(lr){2-2} \cmidrule(lr){3-5}
    {} & \name-0 & 6.5h & -- & 2.8h  \\
    {} & \name-1 & 9.8h & -- & 4.4h  \\
    {} & \name-2 & 13.9h & -- & 6.1h  \\
    {} & \name-3 & 16.0h & -- & 7.9h \\
    \midrule
    \multirow{5}{*}{ImageNet-22K} & input size & $224 \times 224$ & $224 \times 224$ & $384 \times 384$ \\
    \cmidrule(lr){2-2} \cmidrule(lr){3-5}
    {} & \name-0 & 20.1h & 0.9h & 2.5h \\
    {} & \name-1 & 30.0h & 1.3h & 3.9h \\
    {} & \name-2 & 42.6h & 1.8h & 5.4h \\
    {} & \name-3 & 49.2h & 2.2h & 7.0h \\
    \end{tabular}}
\end{table}

\begin{table}[!ht]
    \centering
    \caption{\textbf{ImageNet peak training memory} of MOAT models. The input size is $224 \times 224$.
    }
    \label{tab:peak_training_memory}
    \scalebox{0.8}{
    \begin{tabular}{c|c|c|c|c}
    \multirow{2}{*}{model} & \multicolumn{4}{c}{training statistics} \\
    \cmidrule(lr){2-5}
    {} & total batch size & num. of TPUv4 cores & batch size per core & peak memory per core (MB) \\
    \toprule
    \name-0 & 4096 & 16 & 256 & 19155 \\
    \name-1 & 4096 & 16 & 256 & 26170 \\
    \name-2 & 4096 & 16 & 256 & 26662 \\
    \name-3 & 4096 & 32 & 128 & 26260 \\
    \end{tabular}}
\end{table}

\begin{table}[!ht]
    \centering
    \caption{\textbf{ImageNet throughput measurement} of MOAT models. We re-implement MOAT with the popular ``timm''~\citep{rw2019timm} library in PyTorch, and measure the throughput on an Nvidia V100 GPU, following the same settings as DeiT~\citep{touvron2021training}, Swin~\citep{liu2021swin}, and ConvNeXt~\citep{liu2022convnet}.
    }
    \label{tab:throughput}
    \scalebox{0.8}{
    \begin{tabular}{cc|cc|cc|cc}
    \multicolumn{2}{c}{input size} & \multicolumn{2}{c}{$224 \times 224$} 
    & \multicolumn{2}{c}{$384 \times 384$} & \multicolumn{2}{c}{$512 \times 512$} \\
    \cmidrule(lr){1-2} \cmidrule(lr){3-8}
    \multirow{2}{*}{model} & \multirow{1}{*}{params} & FLOPs & throughput & FLOPs & throughput & FLOPs & throughput  \\
    {} & \multirow{1}{*}{(M)} & (B) & {(images/sec)} & (B) & {(images/sec)} & (B) & {(images/sec)}  \\
    \toprule
    \name-0 & 27.8 & 5.7 & 536 & 18.2 & 155 & -- & -- \\
    \name-1 & 41.6 & 9.1 & 339 & 29.6 & 91 & 58.7 & 41 \\
    \name-2 & 73.4 & 17.2 & 209 & 54.3 & 58 & 104.6 & 27 \\
    \name-3 & 190.0 & 44.9 & 89 & 141.2 & 23 & 271.0 & 9 \\
    \name-4 & 483.2 & -- & -- & -- & -- & 648.5 & 4 \\
    \end{tabular}}
\end{table}

\subsection{Limitations}
\label{sec:limitations_appendix}
Currently, the scaling rule of MOAT model variants are hand-designed. We, therefore,
expect the architecture could be further improved by the breakthroughs in neural architecture search
or network pruning (attaining faster inference speed while maintaining a similar accuracy).

\end{document}